\documentclass[review]{elsarticle}

\usepackage{times}
\usepackage{epsfig}
\usepackage{graphicx}
\usepackage{amsmath}
\usepackage{amssymb}

\usepackage{caption}
\usepackage{subcaption}

\usepackage{rotating}

\usepackage{color, colortbl}

\usepackage{makecell}

\definecolor{Gray}{gray}{0.9}

\newcolumntype{P}[1]{>{\centering\arraybackslash}p{#1}}

\usepackage{lineno,hyperref}
\modulolinenumbers[5]

\journal{Journal of \LaTeX\ Templates}

\begin{document}

\begin{frontmatter}

\title{
Cross-Sensor Periocular Biometrics 
in a Global Pandemic: Comparative Benchmark and
Novel Multialgorithmic Approach
}

\author[HH]{Fernando Alonso-Fernandez\corref{cor}}
\cortext[cor]{Corresponding author}
\ead{feralo@hh.se}
\author[NTNU]{Kiran B. Raja}\ead{kiran.raja@ntnu.no}
\author[NTNU]{R. Raghavendra}\ead{raghavendra.ramachandra@ntnu.no}
\author[NTNU]{Christoph Busch}\ead{christoph.busch@ntnu.no}
\author[HH]{Josef Bigun}\ead{josef.bigun@hh.se}
\author[UAM]{Ruben Vera-Rodriguez}\ead{ruben.vera@uam.es}
\author[UAM]{Julian Fierrez}\ead{julian.fierrez@uam.es}

%
%
%

\address[HH]{School of Information Technology, Halmstad University, Sweden}

\address[NTNU]{Norwegian University of Science and Technology, Gj\o{}vik, Norway}

\address[UAM]{School of Engineering, Universidad Autonoma de Madrid, Spain}

\begin{abstract}
The massive availability of cameras and personal devices results in
a wide variability between imaging conditions, producing large
intra-class variations and a significant performance drop if images
from heterogeneous environments are compared for person recognition
purposes.
However, as biometric solutions are extensively deployed,
it will be
common to replace acquisition hardware as it is damaged or newer
designs appear or to exchange information between agencies or
applications operating in different environments.
Furthermore, variations in imaging spectral bands can also occur.
For example, face images are typically acquired in the visible (VIS)
spectrum, while iris images are usually captured in the near-infrared (NIR)
spectrum. However, cross-spectrum comparison may be needed if, for
example, a face image obtained from a surveillance camera needs to be
compared against a legacy database of iris imagery.
Here, we propose a multialgorithmic approach to cope with
periocular 
images captured with different sensors.
With face masks in the front line to fight against the COVID-19 pandemic,
periocular recognition is regaining popularity since it is the only region of the face that remains visible.
%
As a solution to the mentioned cross-sensor issues, we integrate different biometric comparators
using a score  fusion scheme based on linear logistic regression
This approach is trained to improve the discriminating ability and, at the same time, to encourage that fused scores are represented by log-likelihood ratios. This allows easy
interpretation of output scores
and the use of Bayes thresholds for optimal decision-making
since scores from different comparators are in the same probabilistic range.
%
We evaluate our approach in the context of the 1st
Cross-Spectral Iris/Periocular Competition, whose aim was to compare
person recognition approaches when periocular data from visible and
near-infrared images is matched. The proposed fusion approach
achieves reductions in the error rates of up to 30-40\%
in cross-spectral NIR-VIS comparisons with respect to the best individual system,
leading to an EER of 0.2\% and a FRR of just 0.47\% at FAR=0.01\%. It also
represents the best overall approach of the mentioned competition.
Experiments are also reported with a database of VIS images from two
different smartphones as well, achieving even bigger relative improvements
and similar performance numbers.
We also discuss the proposed approach from the point of view of
template size and computation times, with the most computationally
heavy comparator playing an important role in the results.
Lastly, the proposed method is shown to outperform other
popular fusion approaches in multibiometrics, such as the
average of scores, Support Vector Machines, or Random Forest.
\end{abstract}


\end{frontmatter}


\section{Introduction}
%
%
%
%
Periocular biometrics has gained attention during the last years as an
independent modality for person recognition \cite{[Alonso16],[Nigam15]}
after concerns of the
performance of face or iris modality under non-ideal or uncooperative conditions
\cite{[Proenca18_IntelSys_Trends],[Gonzalez-Sosa18_TIFS_SoftWild]}.
The mandatory use of face masks due to the COVID-19 pandemic has produced that, even in cooperative settings, face recognition systems are presented with occluded faces where the periocular region is often the only visible area.
This face occlusion comes with a reduction in facial information that may be significant for recognition \cite{Tome13_FSI_FacialRegions,Tome15_JFS_FusionFaceRegions}. To what extent this information reduction is detrimental for face recognition is yet something largely unexplored.
In practice, recent studies have shown that commercial face recognition engines, even in cooperative settings, struggle with persons wearing face masks \cite{[Ngan20NISTmasksreport]},
driving vendors to include capabilities for recognition of masked faces in their products \cite{[Klare20rankonemasks]}.
%
In parallel, hygiene concerns are triggering fears against the use of contact-based biometric solutions such as fingerprints \cite{[btt20covidFPshut]}.

\begin{figure}[htb]
\centering
\includegraphics[width=0.45\textwidth]{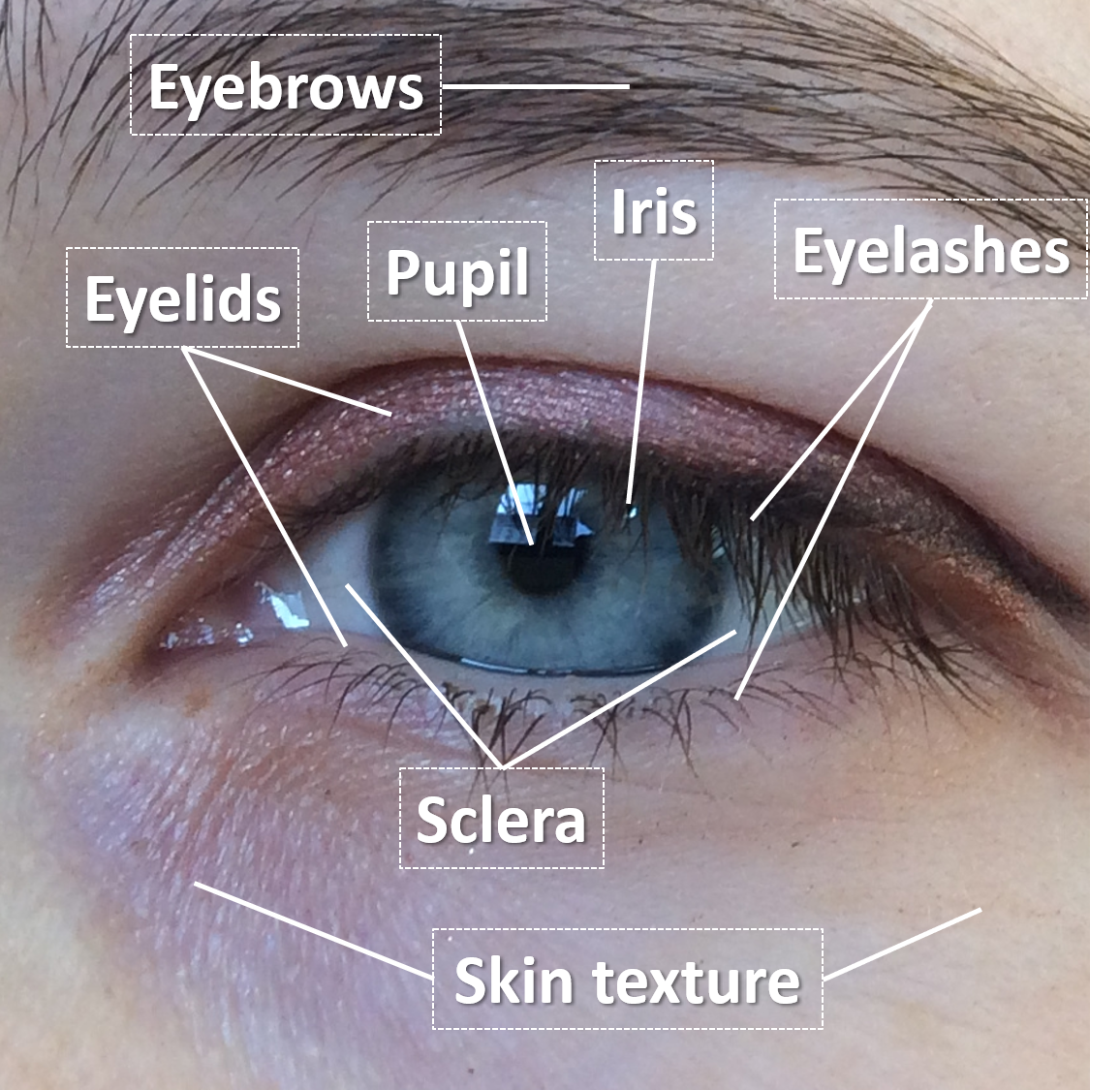}
\caption{Eye image labeled with some parts of the ocular region.}
\label{fig:ocular-parts}
\end{figure}

According to the Merriam-Webster dictionary, the medical definition of
``periocular'' is
``surrounding the eyeball but within the orbit''.
From a forensic/biometric application perspective,
our goal is to improve the recognition performance by
using information extracted from the face region in the
immediate vicinity of the eye, including the sclera, eyelids,
eyelashes, eyebrows and the surrounding skin
(Figure~\ref{fig:ocular-parts}).
%
This information may include textural descriptors,
but also the shape of the eyebrows
or eyelids,
or colour information \cite{[Alonso16]}.
With a surprising high
discrimination ability, the resulting modality is the ocular one requiring
the least constrained acquisition.
It is sufficiently visible over a wide range of
distances, even under partial face occlusion (close distance) or low-resolution iris (long distance), facilitating increased performance
in unconstrained or uncooperative scenarios.
It also avoids the need
for iris segmentation, an issue in difficult images
\cite{[Jillela13ch14]}.
The COVID-19 outbreak has imposed the necessity of dealing with partially occluded faces even in cooperative applications in security, healthcare, border control or education.
Another advantage in the context of the current global pandemic is that the periocular
region appears in iris and face images, so it can be easily obtained
with existing setups for face and iris.

The ocular region consists of several organs such as the cornea, pupil, iris, sclera, lens, retina, optical nerve, and periocular region.
Some of them are shown in Figure~\ref{fig:ocular-parts}.
Among these, iris, sclera, retina and periocular have been studied as biometric modalities \cite{[Nigam15]}.
The significant progress of ocular biometrics in the last decade has been primarily
%
%
due to efforts in iris recognition since the late 80s,
resulting in large-scale deployments \cite{[Jain16]}.
Iris provides
very high accuracy with near-infrared (NIR) illumination and controlled,
close-up acquisition. However, deployment
to non-controlled environments is not yet mature due to the impact of low resolution, variable illumination, or off-angle views, which makes very difficult to locate and segment the iris \cite{[Jillela13ch14]}. Even if the latter can be achieved, the quality of the resulting iris image might not be sufficient for accurate recognition either \cite{[Park11]}.
%
%
The feasibility of vasculature of the sclera as a biometric modality (sometimes simply referred to as sclera) has also been established by several studies \cite{[Rattani17soaOcularVIS]},
although its acquisition in non-controlled environments poses the same problems as the iris modality.
The vasculature of the retina is also very discriminative, and the retina is regarded as the most secure biometric modality due to being extremely difficult to spoof. However, its acquisition is very invasive, requiring high user cooperation and specialized optical devices.
%
%
%

\begin{figure}[htb]
\centering
\includegraphics[width=0.45\textwidth]{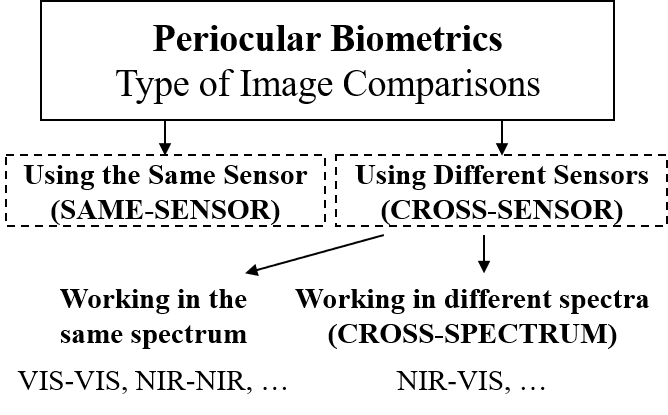}
\caption{Sensor interoperability in periocular biometrics.}
\label{fig:biometrics-interoperability}
\end{figure}

In this context, periocular has rapidly evolved as a very popular modality for unconstrained biometrics \cite{[Nigam15],[Alonso16],[Rattani17soaOcularVIS]}, and recently due to the use of face masks even in constrained settings \cite{[Ngan20NISTmasksreport]}.
The term periocular is used loosely in the literature to refer to the externally visible region of the face that surrounds the eye socket. Therefore, images of the whole eye, such as the one in Figure~\ref{fig:ocular-parts}, are employed as input \cite{[Rattani17soaOcularVIS]}.
While the iris, sclera and other elements are present in such images, they are not explicitly used in isolation.
It may be that the iris texture or the vasculature of the sclera cannot be reliably obtained either to be used as stand-alone modalities \cite{[Park11]}.
Some works even suggest that with visible light data, recognition performance is improved if components inside the ocular globe (iris and sclera) are discarded \cite{[Proenca18periocular_noirissclera]}.
The fast-growing uptake of face technologies in social networks and
smartphones, as well as the widespread use of surveillance cameras or face masks,
has arguably increased the interest in periocular biometrics, especially
in the visible (VIS) range.
In such scenarios, samples captured with different sensors are to be
compared if, for example, users are allowed to use their own
acquisition devices, leading to a \textit{cross-sensor} comparison in the
same spectrum (VIS-VIS in this case).
%
Unfortunately, this massive availability of cameras results in
heterogeneous quality between images \cite{[Alonso12a]}, which is known to decrease
recognition performance significantly \cite{[Jain16]}.
%
These sensor \textit{interoperability} issues also arise when a biometric sensor
is replaced with a newer one without reacquiring the corresponding
template, thus forcing biometric samples from different sensors to
co-exist.
Sensors may also operate in a range other than VIS,
such as NIR, leading to cross-sensor NIR-NIR
comparisons, e.g. \cite{[Xiao12]}.
In addition, iris images are largely acquired
beyond the visible spectrum \cite{[Moreno09]},
mainly using NIR illumination,
but there are several scenarios in which it may be necessary to
compare them with periocular images in the VIS range, leading in
this case to a \textit{cross-sensor}
comparison in different spectra
(NIR-VIS in this case),
also known as \textit{cross-spectral} comparison.
This happens, for example, in law enforcement
scenarios where the only available image of a suspect is obtained
with a surveillance camera in the VIS range,
but the reference database contains images in the NIR range
\cite{[Jillela14],[Tome15FSI_FacialSoftBio]}.
These interoperability problems, if not properly addressed,
can affect the recognition performance dramatically.
Unfortunately, widespread deployment of biometric technologies will
inevitably cause the replacement of hardware parts as they are
damaged, or newer designs appear.
Another application case is the exchange of information among
agencies or applications which employ different technological
solutions or whose data is captured in heterogeneous environments.
The different types of image comparisons mentioned, based on the spectrum in which they have been acquired, are summarized in Figure~\ref{fig:biometrics-interoperability}.



Accordingly, to counteract the reduction in recognition performance that is usually observed when comparing data from different sensors, we propose to
combine the output of different periocular comparators at the score
level, referred to as multialgorithm fusion (in contrast to
multimodal fusion, which combines information from different
modalities) \cite{[Fierrez18],[Lumini17]}. The consolidation of identity evidence
from heterogeneous comparators
(also called experts, feature extraction techniques, or systems in the present paper)
is known to
increase recognition performance, because the different sources can
compensate for the limitations of the others
\cite{[Fierrez18],[Singh19inffus]}. Integration at the
score level is the most common approach because it only needs the
output scores of the different comparators, greatly facilitating the
integration.
With this motivation, we employ a multialgorithm fusion approach
to cope with periocular images from different sensors
which integrates scores from different comparators.
It follows a probabilistic fusion approach
based on linear logistic regression \cite{[Alonso10]}, in which
the output scores of multiple systems are combined
to produce a log-likelihood
ratio according to a probabilistic Bayesian framework.
This allows easy
interpretation 
of output scores 
and the use of Bayes thresholds for optimal decision-making.
This fusion scheme is compared with a set of simple and
trained fusion rules widely employed in multibiometrics
based on the
arithmetic average of normalized scores \cite{[Jain05]},
Support Vector Machines \cite{[Gutschoven00svmfusion]},
and Random Forest \cite{[Ma05rffusion]}.

The fusion approach based on linear logistic regression
served as an inspiration to our submission to the 1$^{st}$ Cross-Spectral
Iris/Periocular Competition (Cross-Eyed 2016)
\cite{[sequeira16crosseyed]}, with an outstanding recognition accuracy:
Equal Error Rate (EER) of 0.29\%, and False Rejection Rate (FRR) of 0\%
at a False Acceptance Rate (FAR) of 0.01\%,
resulting in the best overall competing submission.
This competition was aimed at evaluating the capability of
periocular recognition algorithms to compare
visible and near-infrared images (NIR-VIS).
%
%
In the present paper,
we also carry out cross-sensor experiments with periocular
images in the visible range only (VIS-VIS),
but with two different sensors.
For this purpose, we employ
a database 
captured with two smartphones
\cite{[Raja14b]}, demonstrating the benefits of the proposed
approach to smartphone-based biometrics as well.
%
%
%

The rest of the paper is organized as follows.
This introduction is completed with a description of
the paper contributions. A summary of related works in periocular biometrics is given in Section~\ref{sect:soa}.
Section~\ref{sect:matchers}
then describes the periocular comparators employed.
The score fusion methods evaluated are described in Section~\ref{sect:fusion-methods}.
Recognition experiments using images in different spectra (cross-spectral NIR-VIS)
and in the visible spectrum (cross-sensor VIS-VIS) are
%
described in Sections~\ref{sect:cross-eyed} and
\ref{sect:vssiris}, respectively, including the databases, protocol
used, results of the individual comparators, and fusion experiments.
Finally, conclusions are given in Section~\ref{sect:conclusion}.

\begin{figure}[htb]
\centering
\includegraphics[width=0.45\textwidth]{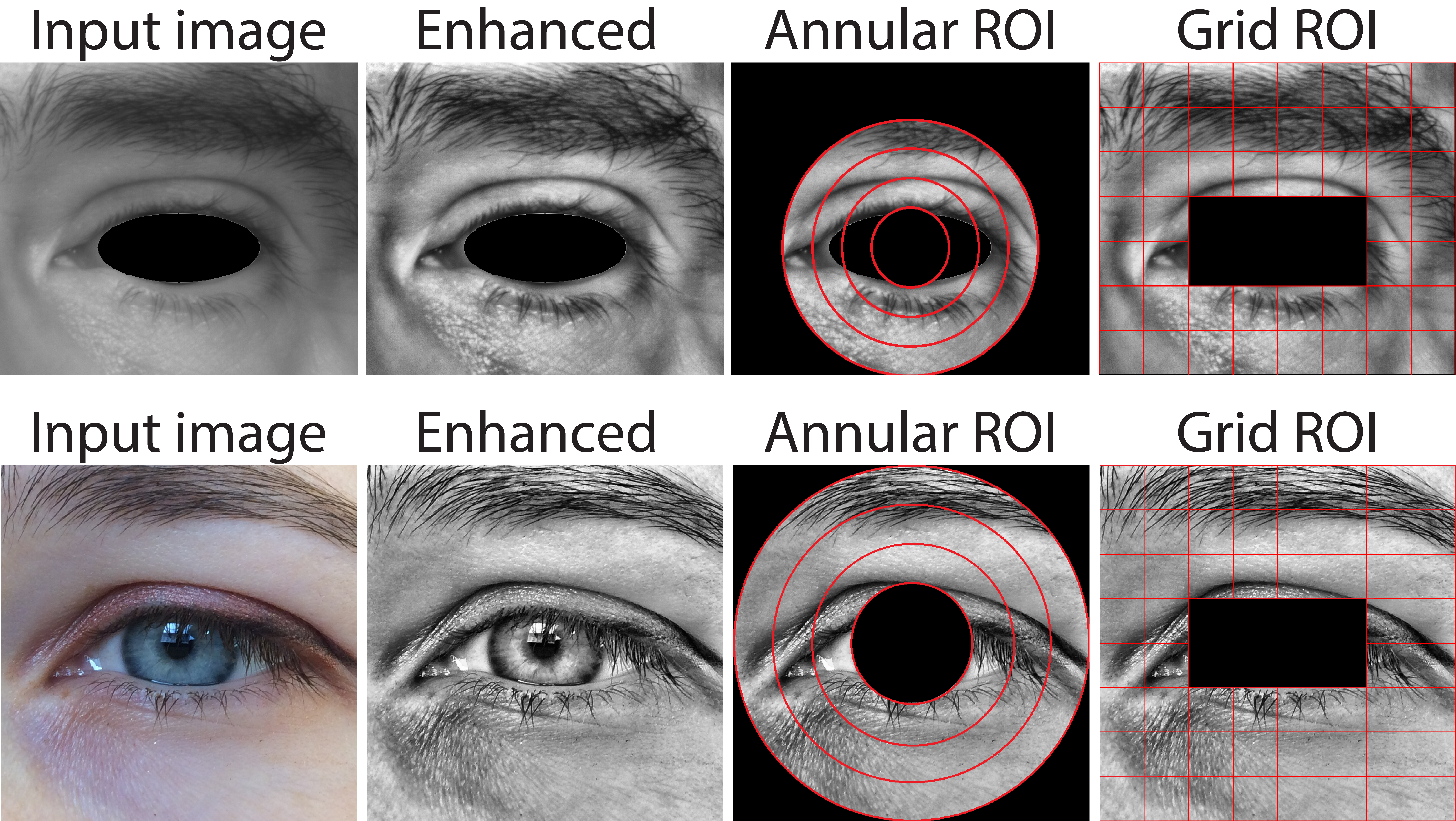}
\caption{Example images from Cross-Eyed (top row) and VSSIRIS
(bottom row) databases. First column: input image. Second: after
applying CLAHE (see Sect.~\ref{sect:cross-eyed-dbprotocol}).
Third and fourth: ROI of the different biometric comparators (see Sect.~\ref{sect:matchers}).}
\label{fig:img-ROI}
\end{figure}

\subsection{Contributions}

\begin{figure}[t]
\centering
\includegraphics[width=0.45\textwidth]{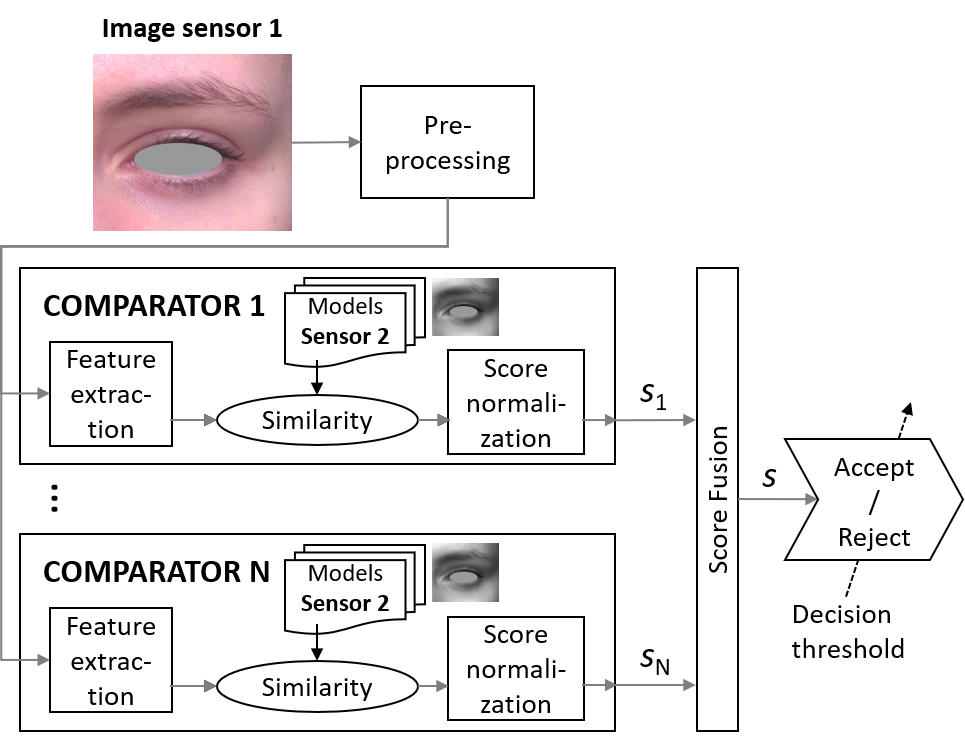}
\caption{Architecture of the proposed fusion strategy.}
\label{fig:system_model}
\end{figure}

\begin{figure}[htb]
\centering
\includegraphics[width=0.45\textwidth]{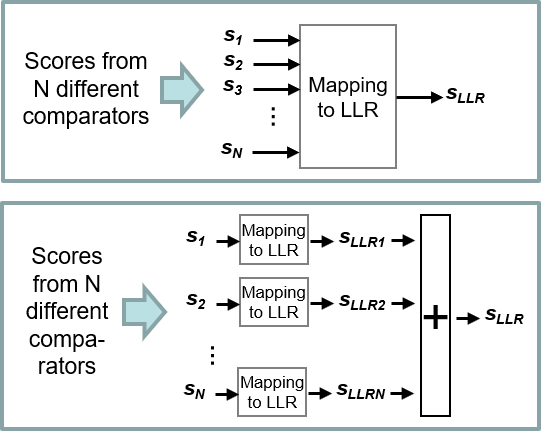}
\caption{Strategies to convert scores from multiple subsystems to a log-likelihood ratio (LLR). Top: one single mapping function is trained to convert multiple scores into a single LLR. Bottom: several mapping functions are trained to convert the score of each subsystem into a LLR. The sum of LLRs from different subsystems also results in a LLR. See the text for details.}
\label{fig:fusion_model}
\end{figure}

The contribution of this paper to the state of the art is thus as
follows.
First, we summarize related works in
periocular biometrics using images from different sensors.
Second, we evaluate nine periocular recognition
comparators under the frameworks of different spectra (NIR-VIS) and
same spectrum (VIS-VIS) recognition. The Reading Cross-Spectral
Iris/Periocular Dataset (Cross-Eyed) \cite{[sequeira16crosseyed]} and the Visible
Spectrum Smartphone Iris (VSSIRIS) \cite{[Raja14b]} databases are
respectively used for this purpose. 
We employ the three most widely
used comparators in periocular research, 
which are used as a baseline in many studies \cite{[Alonso16]}:
Histogram of Oriented Gradients (HOG) \cite{[Dalal05]}, Local Binary Patterns (LBP) \cite{[Ojala02]}, and
Scale-Invariant Feature Transform (SIFT) key-points \cite{[Lowe04]}.
Three other periocular comparators, proposed and published previously by the authors, are based on Symmetry Descriptors \cite{[Alonso16a]},
Gabor features \cite{[Alonso15]},
and Steerable Pyramidal Phase Features \cite{[Raja17]}.
The last three comparators use feature vectors extracted by three Convolutional Neural Networks:
VGG-Face \cite{[Parkhi15]},
which has been trained for classifying faces, so the periocular region appears in the training data, and the very-deep Resnet101 \cite{[He16]} and Densenet201 \cite{[Huang17]} networks.
%
%
Two example images from the two databases employed are shown in Figure~\ref{fig:img-ROI} (first column). The second column shows the two images after applying Contrast Limited Adaptive Histogram
Equalization (CLAHE) \cite{[Zuiderveld94clahe]}, whereas the last two columns show the regions of interest (ROI) used by the different comparators.
%
%
The comparators are evaluated 
both in terms of performance, template size
and computation times.
In a previous study \cite{[Alonso17b_eusipco_busch]}, we presented preliminary
results with the VSSIRIS database using a subset
of the mentioned comparators \cite{[Park11],[Alonso16a],[Alonso15]},
which are extended in the present paper with additional experiments
using new comparators \cite{[Raja17],[Parkhi15],[He16],[Huang17]}
and the mentioned Cross-Eyed database.
Third, we describe our multialgorithm fusion architecture for
periocular recognition using images from different sensors (Figure~\ref{fig:system_model}).
The input to a biometric comparator is usually a pair of biometric samples,
and the output is, in general, a similarity score $s$.
A larger score favours the hypothesis that the two samples
come from the same subject (target or client hypothesis),
whereas a smaller
score supports the opposite (non-target or impostor hypothesis).
However, if we consider a single isolated score
from a biometric comparator (say a similarity score of $s$=1),
it is in general not possible to determine
which is the hypothesis the score supports the most, unless
we know the distributions of target or non-target scores.
Moreover, since the scores
output by the various comparators are heterogeneous, score
normalization is needed to transform these scores into a common
domain prior to the fusion process
\cite{[Fierrez18]}.
We solve these problems by
linear logistic regression fusion
\cite{[pigeon00],[brummer07fusion]}, a trained classification
approach in which 
scores of the individual comparators are combined to obtain
a log-likelihood
ratio. This is the logarithm of the ratio between the likelihood
that input signals were originated by the same subject and the
likelihood that input signals were not originated by the same
subject. This form of output is comparator-independent in the sense
that this log-likelihood-ratio output can theoretically be used to
make optimal (Bayes) decisions.
%
%
To convert scores from different comparators into a log-likelihood ratio, we evaluate two possibilities (Figure~\ref{fig:fusion_model}).
In the first one (top part), the mapping function uses as input the scores of all comparators, producing a single log-likelihood ratio as output.
In the second one (bottom), several mapping functions are trained (one per comparator), so one log-likelihood ratio per comparator is obtained.
%
%
Under independence
assumptions (as in the case of comparators based on different feature extraction methods),
the sum of
log-likelihood ratios results in another log-likelihood ratio
\cite{[DudaHartStork]}. Therefore, in the second case, the outputs of the different mapping functions are just summed.
The latter provides a simple fusion framework that allows obtaining  a single log-likelihood ratio
by simply summing the (mapped) score given by each available comparator.
This would allow coping with missing modalities \cite{[Poh09]} since the output still would be a log-likelihood ratio regardless of the number of systems combined.
%
%
This fusion approach has been previously applied successfully to
cross-sensor comparison in the face and fingerprint modalities
\cite{[Alonso10]}, achieving excellent results in other competition
benchmarks as well \cite{[Poh09]}.
Fourth, we compare this fusion approach with a set of
simple and
trained score fusion rules based on the
arithmetic average of normalized scores \cite{[Jain05]},
Support Vector Machines \cite{[Gutschoven00svmfusion]},
and Random Forest \cite{[Ma05rffusion]}.
These fusion approaches are very popular in the literature,
having demonstrated to give good results in biometric authentication \cite{[Fierrez06Tesis],[Fierrez18]}.
Fifth, in our experiments,
conducted according to the
1st Cross-Spectral Iris/Periocular Competition (Cross-Eyed 2016) protocol \cite{[sequeira16crosseyed]},
reductions of up to 29/47\% in EER/FRR error
rates (with respect to the best individual system)
are obtained by fusion under NIR-VIS comparison,
resulting in a cross-spectral EER of 0.2\%, and a FRR @ FAR=0.01\% of
just 0.47\%. 
Regarding cross-sensor VIS-VIS
smartphone recognition, the reductions in error
rates achieved are 85/93\% in EER/FRR, respectively, with
corresponding cross-sensor error values of 0.3\% (EER) and 0.3\%
(FRR).

\setlength{\tabcolsep}{0pt}

\begin{table}[htb]
\tiny

\begin{center}
\begin{tabular}{|c|c|c|c||c|c|c|c|c|c|c|}

\multicolumn{11}{c}{} \\ \cline{5-11}

\multicolumn{4}{c}{} &
\multicolumn{7}{|c|}{\textbf{Best accuracy}}\\ \hline

 &  &  & \textbf{People/} &  \textbf{} &  &  &
\textbf{GAR @} &
\textbf{GAR @} &
\textbf{GAR @} & \\

\textbf{Ref.} &
\textbf{Features} &
\textbf{Database} &
\textbf{Images} &
\textbf{Comparison} &
\textbf{$\#$ Eyes} &
 \textbf{EER} &
\textbf{1\%FAR} &
\textbf{0.1\%FAR} &
\textbf{0.01\%FAR} &
\textbf{Rank-1}
\\ \hline

\multicolumn{11}{c}{} \\

\multicolumn{11}{c}{\textbf{Cross-sensor comparisons in the visible range (VIS-VIS)}} \\ \hline

\cite{[Santos14]} & LBP, HOG, SIFT, ULBP, GIST & CSIP &  50/2004 & VIS-VIS & single
& 15.5\% & - & - & - & - \\ \hline \hline


\cite{[Raja16]} & LD+STFT & MICHE I &  50/n-a & VIS-VIS & single
& 6.38-8.33\% & - & - & - & - \\
\hline \hline

\cite{[Kandaswamy17]} & GMM-UBM, SV-SDA, CNN & CSIP &  50/2004 & VIS-VIS & single
& - & - & - & - & 83.6-93.3\% \\ \hline  \hline


\multicolumn{2}{|c|}{\textbf{this work:} 9 comparators} & VSSIRIS &  56/560 & VIS-VIS & single
& 0.3\% & - & - & 99.7\% & - \\ \hline

\multicolumn{11}{c}{} \\

\multicolumn{11}{c}{\textbf{Cross-sensor comparisons in the near-infrared range (NIR-NIR)}} \\ \hline

\cite{[Xiao12]} & OM & own &  300/9000 & NIR-NIR & single
& 20-28\% & - & - & - & - \\ \hline

\multicolumn{11}{c}{} \\

\multicolumn{11}{c}{\textbf{Cross-sensor comparisons across different spectra (cross-spectrum)}} \\ \hline

\cite{[Jillela14]} & LBP, NGC, JDSR & own &  704/1358 & VIS-NIR & single
& 23\% & - & - & - & - \\ \hline \hline

\cite{[Sharma14]} & PHOG & IIITD-IMP &  62/1240 & VIS-NIR & single/both
& - & 38.36/47.08\% & - & - & - \\ \cline{5-11}

& & & &  VIS-night & single/both
& - & 63.81/71.93\% & - & - & - \\ \cline{5-11}

& & & &  NIR-night & single/both
& - & 40.36/48.21\% & - & - & - \\ \hline \hline

\cite{[Cao16]} & Gabor+ & Pre-Tinders &  48/576 & VIS-SWIR 1.5/50/106m & single
& 7.32/24.87/31.18\% & - & - & - & 68.75/33.33/31.94\% \\ \cline{5-11}

& WLD/LBP/HOG & Tinders  &  48/1255 & VIS-NIR 1.5/50/106m & single
& 4.42/25.71/39.01\% & - & - & - & 70.31/38.54/10.76\% \\ \cline{5-11}

& & PCSO & 1000/3000 &  VIS-MWIR 1.5m & single
& 30.46\% & - & - & - & 5.58\% \\ \cline{5-11}

& & Q-FIRE & 82/431 &  VIS-LWIR 2m & single
& 39.06\% & - & - & - & 8.09\% \\ \hline \hline

\cite{[Ramaiah16]} & MRF+ & IIITD IMP &  62/1240 & VIS-NIR & single
& - & - & 15.93-18.35\% & - & - \\ \cline{3-11}

& TPLBP/FPLB & PolyU &  209/12540 & VIS-NIR & single
& 19.8-32.5\% & - & 45.4-73.2\% & - & - \\
\hline \hline

\cite{[Behera17]} & DOG+LBP/HOG & IIITD-IMP &  62/1240 & VIS-NIR & single/both
& 43.85/45.29\% & - & 24.97/25.03\% & - & - \\ \cline{3-11}

 & & PolyU &  209/12540 & VIS-NIR & single/both
& 18.79/13.87\% & - & 73.12/83.12\% & - & - \\  \cline{3-11}

 & & Cross-Eyed &  120/3840 & VIS-NIR & single/both
& 15.11/10.36\% & - & 80.03/89.27\% & - & - \\ \hline \hline

\cite{[Vetrekar18]} & HOG, GIST, LG, BSIF  & own &  52/4160 & 8 bands & both
& - & - & - & - & 8.46-91.92\% \\
\hline \hline

\cite{[Hernandez19]} & CNN & IIITD-IMP &  62/1240 & VIS-NIR & single
& 5.19\% & 88.13\% & - & - & - \\ \cline{5-11}

& &  &   & VIS-night & single
& 5.13\% & 88.19\% & - & - & - \\ \cline{5-11}

& &  &   & NIR-night & single
& 10.19\% & 81.55\% & - & - & - \\ \hline \hline

\multicolumn{2}{|c|}{\textbf{this work:} 9 comparators} & Cross-Eyed &  120/3840 & VIS-NIR & single
& 0.2\% & - & - & 99.53\% & - \\ \hline

\end{tabular}
\end{center}
\caption{Overview of existing works in periocular biometrics using images from different sensors. The works of each sub-section are in chronological order. The acronyms of this table are fully defined in the text.
}
\label{tab:SoA-peri-rec}
\end{table}
\normalsize

\setlength{\tabcolsep}{6pt}

\section{Related Works in Periocular Biometrics Using Images from Different Sensors}
\label{sect:soa}

Interoperability between different sensors is an area of high research interest due
to new scenarios arising from the widespread use of biometric technologies, coupled with the availability of multiple sensors and vendor solutions.
A summary of existing works in the literature is given in
Table~\ref{tab:SoA-peri-rec}.
Most of them employ the Genuine Acceptance Rate (GAR) as metric, which is computed as
100-FRR(\%). For this reason, in this subsection, we report GAR values.
However, in the rest of the paper, we will follow the Cross-Eyed protocol
and will report FRR values.

Cross-sensor comparison of images in the visible range (VIS-VIS)
from smartphone sensors is carried out, for example, in
\cite{[Santos14],[Raja16],[Kandaswamy17]},
while the challenge of comparing images from different sensors
in the near-infrared spectrum
(NIR-NIR) has been addressed in \cite{[Xiao12]}.
In the work \cite{[Raja16]}, the authors apply Laplacian
decomposition (LD) of the image coupled with dynamic scale
selection, followed by frequency decomposition via
Short-Term Fourier Transform (STFT).
In the experiments, they employ a subset of 50
periocular instances from the
MICHE I dataset (Mobile Iris Challenge Evaluation I dataset) \cite{[Marsico14]},
captured with the front
and rear cameras of two smartphones
in indoor and outdoor illuminations.
The cross-sensor EER obtained ranges from 6.38 to 8.33\%
for the different combinations of reference and probe cameras.
The authors in
\cite{[Santos14]} use a sensor-specific colour correction
technique, which is estimated by using a colour chart in a dark
acquisition scene that is further illuminated by a standard
illuminant. The authors also carry out a score-level fusion of six
iris and five periocular comparators, which is done by Neural Networks.
The five periocular features include
Local Binary Patterns (LBP), 
Histogram of Oriented Gradients (HOG), 
Scale-Invariant Feature Transform (SIFT) key-points, 
Uniform Local Binary Patterns (ULBP) \cite{[Guo10ULBPs]},
and the perceptual GIST descriptors \cite{[Oliva01]}.
They also presented a new database (CSIP: Cross-Sensor Iris and Periocular), with 2004 periocular images from 50 subjects
captured with four different
smartphones in ten different setups (based on several combinations
involving the use of frontal/rear cameras and flash/no flash).
The best reported periocular performance by fusion of the five available
comparators is EER=15.5\%.
The same database is also employed in \cite{[Kandaswamy17]},
where the authors apply three different methods to solve the cross-sensor
task: Gaussian Mixture Models coupled with Universal Background Models (GMM-UBM),
GMM Supervectors coupled with Stacked Denoising Autoencoders (SV-SDA),
and deep transfer learning with Convolutional Neural Networks (CNN).
They achieve a rank-1 recognition rate of 93.3\% in the best
possible case.
The work \cite{[Xiao12]}, on the other hand, addresses the issue of
cross-sensor recognition in the NIR spectrum.
The authors employ a self-captured database
with 9000 iris images from 600 eyes (300 people) using three
different high-resolution sensors. Sensor interoperability is dealt with
by weighted
fusion of information from multiple directions of Ordinal Measures (OM),
with a reported cross-sensor periocular EER between 20 and 28\%.

Regarding recognition across different spectra (cross-spectral), the work
\cite{[Jillela14]} proposes to compare images of the
periocular region cropped from VIS face images against NIR iris
images. This is because face images are usually captured in the visible range,
while iris images in commercial systems are usually acquired using
near-infrared illumination. They employ three different
comparators based on Local Binary Patterns (LBP),
Normalized Gradient Correlation (NGC), and
Joint Database Sparse Representation (JDSR).
Using a self-captured database with 1358 images of the left eye from 704 subjects,
they report a
cross-spectral EER of 23\% by score-level
fusion of the
three comparators.

In another line of work, surveillance
at night or in harsh environments has prompted interest
in new imaging modalities.
For example, the authors in \cite{[Sharma14]} presented the
IIITD Multispectral Periocular database (IIITD-IMP), with a total of 1240
VIS, NIR and Night Vision images from 62 subjects
(the latter captured with a video
camera in Night Vision mode). To cope with cross-spectral periocular
comparisons, they employ Neural Networks to learn the variabilities
caused by each pair of spectra. The employed comparator
is based on a Pyramid
of Histograms of Oriented Gradients (PHOG) \cite{[Bosch07]}.
They report results for each eye separately and
for the combination of both eyes, obtaining a cross-spectral
GAR of 38-64\% at FAR=1\%
(best of the two eyes), and a
GAR of 47-72\% combining the two eyes.
The use of pre-trained Convolutional Neural Networks (CNN)
as a feature extraction method for NIR-VIS comparison
was recently proposed in \cite{[Hernandez19]}.
Here, the authors identify the layer of the ResNet101
network that provides the best performance on each spectrum.
Then, they train a Neural Network that uses as input
the feature vector of the best respective layers.
Using the IIITD-IMP database,
they report
results considering the left and right eyes of a person
as different users (effectively duplicating the number of
classes).
The obtained cross-spectral accuracy is
EER=5-10\% and GAR=81-88\% at FAR=1\%,
which outperforms any previous study with this database.
The authors in \cite{[Ramaiah16]} employ the IIITD-IMP database,
and a newly presented database,
the Hong Kong Polytechnic University Cross-Spectral Iris Images Database (PolyU),
with 12540 images from 209 subjects.
To carry out NIR-VIS comparison,
they use Markov Random Fields (MRF) combined with two different feature extraction methods,
variants of Local Binary Patterns (LBP), namely FPLBP
(Four-Patch LBP) and TPLBP (Three-Patch LBP).
They report a cross-spectral periocular GAR at FAR=0.1\%
of 16-18\% (IIITD-IMP) and 45-73\% (PolyU).
These two databases, together with the Cross-Eyed database
(with 3840 images in NIR and VIS spectra from 120 subjects)
\cite{[sequeira16crosseyed]},
are used in the work \cite{[Behera17]}.
To normalize the differences in illumination between NIR and VIS images,
they apply Difference of Gaussian (DoG) filtering.
The comparators employed were based on Local Binary Patterns (LBP) and Histogram of Oriented Gradients (HOG) features.
They report results for each eye separately and
for the combination of both eyes.
The IIITD-IMP database gives the worst results, with
a cross-spectral EER of 45\% and a GAR at FAR=0.1\% of only 25\%
(two eyes combined).
The reported accuracy with the other databases is better,
ranging between 10-14\% (EER) and 83-89\% (GAR).
%

%
%

%
Latest advancements have resulted in devices with the ability to see
through fog, rain, at night, and to operate at long ranges.
In the work \cite{[Cao16]}, the authors carry out experiments with
several databases containing images with
different wavelengths, namely VIS, NIR, SWIR (ShortWave Infrared),
MWIR (MiddleWave Infrared), and LWIR (LongWave Infrared).
The images are captured at several stand-off
distances of 1.5 m, 2 m, 50 m, and 105 m.
Feature extraction is done with a bank of Gabor filters, with the
magnitude and phase responses further encoded with three
descriptors: Weber Local Descriptor (WLD) \cite{[Chen10]}, Local
Binary Patterns (LBP), 
and Histogram of Oriented
Gradients (HOG). 
Extensive experiments are done in this work
comparing SWIR, NIR, MWIR and LWIR periocular probes to
a gallery of VIS images.
As expected, accuracy decreased as the standoff distance increases.
Also, the comparison of MWIR or LWIR images to VIS images
shows poor performance, attributable to the fact that
MWIR and LWIR imagery measures the heat of a body,
while visible imagery measures reflected light.
Recently, the work \cite{[Vetrekar18]} presented a new multispectral
database captured in eight bands across the VIS and NIR spectra
(530 to 1000 nm).
A total of 4160 images from 52 subjects
were acquired using a custom-built sensor that captures
periocular images simultaneously in the eight bands.
The comparators evaluated are based on Histogram of Oriented Gradients (HOG),
perceptual descriptors (GIST), Log-Gabor filters (LG), and
Binarized Statistical Image Features (BSIF).
The cross-band accuracy varies greatly depending on the reference
and probe bands, ranging from 8.46\% to 91.92\% rank-1 identification
rate.

\begin{figure}[htb]
\centering
\includegraphics[width=0.45\textwidth]{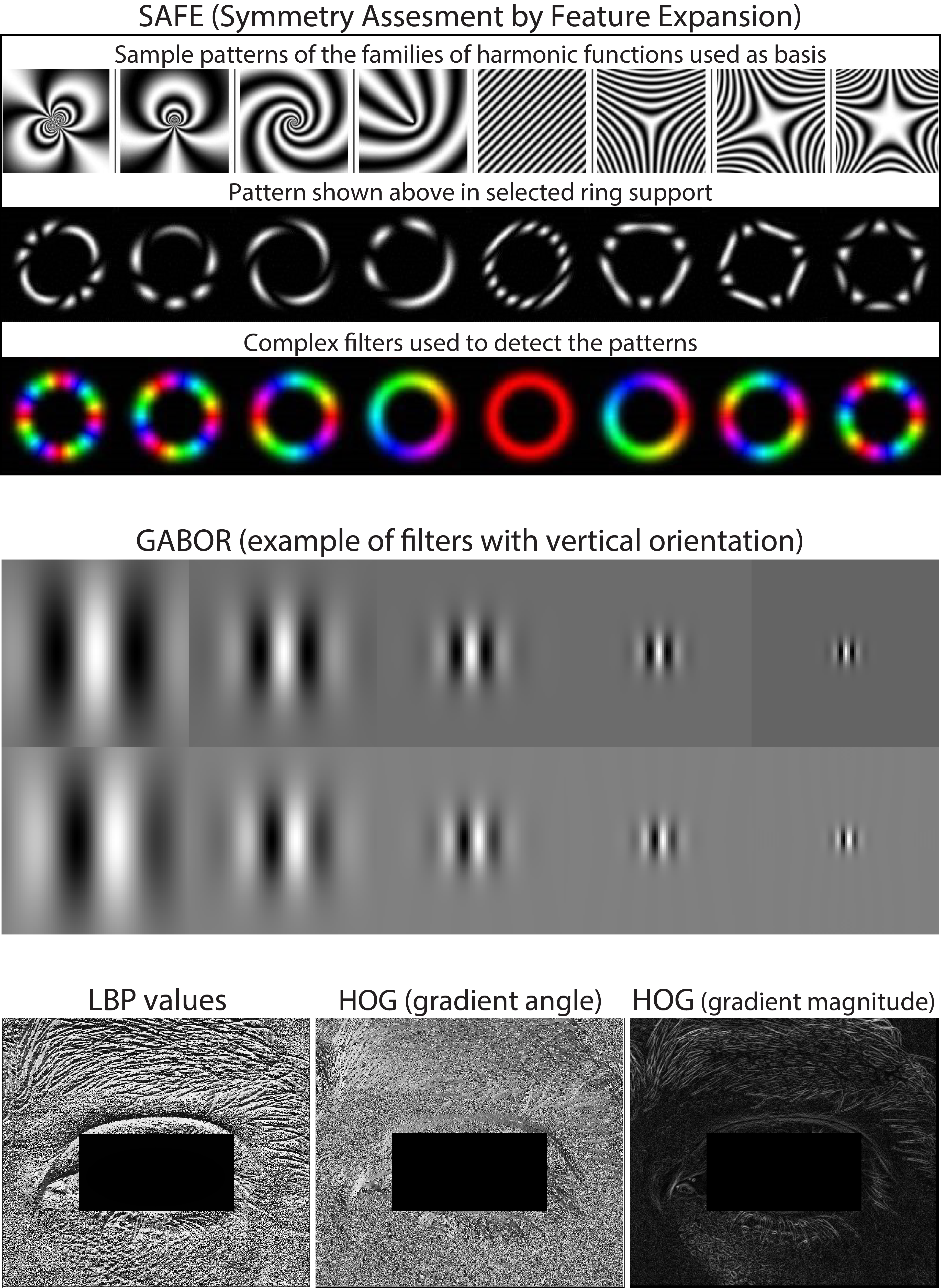}
\caption{Example of some feature extraction methods employed.
\textbf{SAFE comparator.}  Example of symmetric curve families and
complex filters used to detect the patterns.
Hue in
colour images encode the direction, and saturation represents the
complex magnitude.
\textbf{GABOR comparator.} Gabor filters with vertical orientation (top: real part,
bottom: imaginary part). Depicted filters are of size 88$\times$88,
with wavelengths spanning logarithmically the range from 44 (first
column) to 6 pixels (last column).
\textbf{LBP and HOG comparators.} Example of LBP and HOG features of the input image shown in Figure~\ref{fig:img-ROI} (top row).
}
\label{fig:features_safe_gabor_lbp_hog}
\end{figure}



\section{Periocular Comparators}
\label{sect:matchers}

This section describes the biometric comparators used for periocular recognition.
We employ nine different comparators, whose choice is motivated as follows.
Three comparators are based on the most widely used features in periocular research,
which are employed as a baseline in many studies
\cite{[Alonso16]}: Histogram of Oriented Gradients (HOG)
\cite{[Dalal05]}, Local Binary Patterns (LBP) \cite{[Ojala02]}, and
Scale-Invariant Feature Transform (SIFT) key-points \cite{[Lowe04]}.
Other three comparators, available in-house,
have been self-developed by the authors
and published previously with competitive results.
These are based on
Symmetry Descriptors (SAFE) \cite{[Alonso16a]},
Gabor features (GABOR) \cite{[Alonso15]},
and Steerable Pyramidal Phase Features (NTNU) \cite{[Raja17]}.
We also employ three comparators based on deep Convolutional Neural Networks:
the VGG-Face network \cite{[Parkhi15]},
which has been trained for classifying faces
(so the periocular region appears in the training data),
and the two very-deep
Resnet101 \cite{[He16]}
and Densenet201 \cite{[Huang17]}
architectures.

\subsection{Based on Symmetry Patterns (SAFE)}

This comparator employs the Symmetry Assessment by Feature Expansion
(SAFE) descriptor \cite{[Alonso16a]}, which encodes the presence of
various symmetric curve families around image key-points
(Figure~\ref{fig:features_safe_gabor_lbp_hog}, top).
We use the eye centre as the anchor point for feature extraction.
The algorithm starts by extracting the complex orientation map of the
image via symmetry derivatives of Gaussians \cite{[Bigun06]}. 
%
%
We employ $S$=6 different scales in
computing the orientation map, therefore capturing features at
different scales, with standard deviation of each scale given by
$\sigma_s=K^{s-1}\sigma_0$ (with $s=1,2...,S$; $K=2^{1/3}$;
$\sigma_0=1.6$). These parameters have been chosen according to \cite{[Lowe04]}.
%
%
For each scale, we then project $N_f=3$ ring-shaped areas of
different radii around the eye centre onto a space of $N_h=9$
harmonic functions.
We use the result of scalar products of complex harmonic filters
(shown in Figure~\ref{fig:features_safe_gabor_lbp_hog})
with the orientation image to
quantify the amount of presence of different symmetric pattern
families within each annular band.
The resulting complex feature vector is given by an array
of $S \times N_h \times N_f$ elements. 
%
The comparison score $M\in\mathbb{C}$ between
a query $q$ and a test SAFE array $t$
is computed using the triangle inequality as
$
M = \frac{{\left\langle {q,t} \right\rangle }}{{\left\langle {\left| q \right|,\left| t \right|} \right\rangle }}
$.
%
\noindent The argument $\angle M$ represents the angle between the two
arrays (expected to be zero when the symmetry patterns detected
coincide for reference and test feature vectors, and 180$^\circ$
when they are orthogonal), and the confidence is given by $|M|\in[0,1]$. To
include confidence into the angle difference, we use $MS=|M|\cos
\angle M$, with the resulting score $MS\in[-1,1]$.

The annular band of the first ring is set in proportion to the
distance between eye corners (Cross-Eyed database) or to the radius
of the sclera circle (VSSIRIS database), while the band of the last
ring ends at the boundary of the image.
This difference in setting the smallest ring is due to the
ground-truth information available for each database, as
explained later.
However, in setting the origin of the smallest band,
we have tried to ensure that the different annular rings capture approximately
the same relative spatial region in both databases.
The ROI of the SAFE comparator for each database
is shown in Figure~\ref{fig:img-ROI} (third column).
Using the eye corners or the sclera boundary as reference for
the first annular band alleviates the effect of dilation that affects the
pupil, which is more pronounced with visible illumination.
Since the eye corners or the sclera are
not affected by such dilation or by partial occlusion due to eyelids,
they provide a more stable reference \cite{[Padole12]}.

%
%
%

\subsection{Based on Gabor Features (GABOR)}

This comparator is described in \cite{[Alonso15]}, which is based on
the face recognition comparator presented in \cite{[Smeraldi02]}.
The periocular image is decomposed into non-overlapped square
regions (Figure~\ref{fig:img-ROI}, fourth column), and the local
power spectrum is then sampled at the centre of each block by a set
of Gabor filters organized in 5 frequency and 6 orientation
channels.
An example of Gabor filters is shown in
Figure~\ref{fig:features_safe_gabor_lbp_hog}.
%
This sparseness of the sampling grid allows direct Gabor filtering
in the image domain without needing the Fourier transform, with
significant computational savings and feasibility in real-time.
%
%
Gabor responses from all grid points are grouped into a single
complex vector, and the comparison between
two images
is made using the magnitude of complex
values via the $\chi^2$ distance. Prior to the comparison
with magnitude
vectors, they are normalized to a probability distribution (PDF).
The $\chi^2$ distance between a query $q$ and a test vector $t$
is computed as
$
\chi_{qt}^2 = \sum\limits_{n = 1}^{N}{\frac{(p_{q}[n] -
p_{t}[n])^2}{p_{q}[n] + p_{t}[n]}}
$,
where $p$ are entries in the PDF, $n$ is the bin index,
and $N$ is the number of bins in the PDF (dimensionality).
The $\chi^2$ distance, due to the denominator, gives more weight to
low probability regions of the PDF. For this reason, it has been
observed to produce better results than other distances when using
normalized histograms \cite{[Bulacu07]}.
%
%

\subsection{Based on Steerable Pyramidal Phase Features (NTNU)}

Image features from multi-scale pyramids have proven to extract discriminative features in many earlier works
concerned with texture synthesis, texture retrieval, image fusion, and texture classification, among others \cite{Unser2011SteerablePA, do2002rotation, tzagkarakis2006rotation, portilla2000parametric, lyu2009modeling, li2004comparison, el2009novel,su2005steerable}. Inspired by this applicability, we employ steerable pyramidal features for periocular image classification using images from different sensors. Further, observing the nature of textures that are different across spectra (NIR versus VIS), we propose to employ the quantized phase information from the multi-scale pyramid of the image, as explained next.

A steerable pyramid is a translation and rotation invariant transform in a multi-scale, multi-orientation and self-inverting image decomposition into a number of sub-bands \cite{simoncelli1995steerable, freeman1991design, greenspan1994rotation}. The pyramidal decomposition is performed using directional derivative operators of a specific order. The key motivation in using steerable pyramids is to obtain both linear and shift-invariant features in a single operation. Further, they not only provide multi-scale decomposition but also provide the advantages of orthonormal wavelet transforms that are both localized in space and spatial-frequency with aliasing effects \cite{simoncelli1995steerable}. The basis functions of a steerable pyramid are $K$-order directional derivative operators. 
The steerable pyramids come in different scales and $K+1$ orientations.

For a given input image, 
the features of steerable pyramid coefficients can be represented using $S_{(m, \theta)}$, where $m$ represents the scale and $\theta$ represents the orientation. In this work, we generate a steerable pyramid 
with 3 scales ($m \in \left\{ {1,2,3} \right\}$) and
angular coefficients in the range $\theta_1= 0$ to $\theta_{K+1} = 360$,
resulting in a pyramid that covers all directions. The set of sub-band images corresponding to one scale 
can be therefore represented as
$S_m = \{S_{(m, \theta_1)}, \ S_{(m, \theta_2)}, \ldots S_{(m, \theta_{K+1})}\}$.
%
We further note that the textural information represented is different in the NIR and VIS domains. In order to 
obtain domain invariant features, we propose to extract the local phase features \cite{ojansivu2008blur} from each sub-band image $S_{(m, \theta)}$ in a local region $\omega$ in the neighbourhood of $n$ pixels given by
%
$F_{(m, \theta)}(u,x) =  S_{(m, \theta)}(x,y)\omega_R(y - x)\exp\{{-j2\pi U^Ty}\}$,
%
%
where $x,y$ represent the pixel location. 
The local phase response obtained through Fourier coefficients are computed for the frequency points $u_1, u_2, u_3$ and $u_4$, which relate to four points $[a,0]^T, [0,a]^T, [a,a]^T, [a,-a]^T$ such that the phase response $H(u_i) > 0$ \cite{ojansivu2008blur}. The phase information presented in the form of Fourier coefficients is then separated into real and imaginary parts of each component, as given by $[Re\{F\}, Im\{F\}]$, to form a vector $R$ with eight elements.
Next, the elements $R_i$ of $R$ are binarized 
to $Q_i$
by assigning a value of $1$ to components with a response greater than $1$,
and $0$ otherwise. 
%
%
The 
phase information is finally encoded to 
a compact pixel representation $P$ in the $0 - 255$ range by using a simple binary to decimal conversion strategy given by 
%
$P_{(m, \theta)} = \sum_{j = 1}^{8} Q_j \times (2^{(j-1)})$.

This procedure is followed with the different scales and orientations of the selected space. 
All the phase responses $P_{(m, \theta)}$ of the input image
are concatenated into a single vector.
Comparison between feature representations of two images is made using the $\chi^2$ distance.

\subsection{Based on SIFT Key-points (SIFT)}

This comparator is based on the SIFT operator \cite{[Lowe04]}.
SIFT key-points (with dimension 128 per key-point)
are extracted in the annular ROI shown in
Figure~\ref{fig:img-ROI}, third column.
The use of an annular ROI like SAFE is inherited from our previous contribution \cite{[Alonso17b_eusipco_busch]}, but to compare with other systems that employ the entire input image (Figure~\ref{fig:img-ROI}, fourth column), we report experiments with the latter as well.
The recognition metric between two images is the number of
paired key-points, normalized by the minimum number of detected
key-points in the two images being compared.
We use a free C++ implementation of the SIFT
algorithm\footnote{http://vision.ucla.edu/$\sim$vedaldi/code/sift/assets/sift/index.html},
with the adaptations described in \cite{[Alonso09]}. Particularly,
it includes a post-processing step to remove spurious pairings
using geometric constraints, so pairs whose
orientation and length differ substantially from the predominant
orientation and length are removed.


\subsection{Based on Local Binary Patterns (LBP) and Histogram
of Oriented Gradients (HOG)}

Together with SIFT key-points, LBP \cite{[Ojala02]} and HOG
\cite{[Dalal05]} have been the most widely used descriptors in
periocular research \cite{[Alonso16]}.
An example of LBP and HOG features is shown in
Figure~\ref{fig:features_safe_gabor_lbp_hog}, bottom.
The periocular image is decomposed into non-overlapped regions, as
with the Gabor comparator (Figure~\ref{fig:img-ROI}, fourth column).
%
%
Then, HOG and LBP features are extracted from each block. Both HOG
and LBP are quantized into 8 different values to construct an 8 bins
histogram per block. Histograms from each block are then normalized
to account for local illumination and contrast variations and
finally concatenated to build a single descriptor of the whole
periocular region.
%
%
Image comparison with HOG and LBP can be made by simple distance measures.
Euclidean distance is usually used for this purpose \cite{[Park11]},
but here we employ the $\chi^2$ distance for the same reasons as
with the Gabor comparator.
%

\subsection{Based on Deep Convolutional Neural Networks (VGG-Face, Resnet101, Densenet201)}

Inspired by the works \cite{[Nguyen18],[Hernandez18],[Hernandez19]} in iris and periocular biometrics,
we leverage the power of existing architectures pre-trained with
millions of images to classify hundreds of thousands of object categories\footnote{ImageNet. http://www.image-net.org}.
They have proven to
be successful in very large recognition tasks apart from the detection and
classification tasks for which they were designed \cite{[Razavian14]}.
%
%

%
Here, we employ the VGG-Face \cite{[Parkhi15]} and the very deep
Resnet101 \cite{[He16]}
and Densenet201 \cite{[Huang17]}
architectures.
VGG-Face is based on the VGG-Very-Deep-16 CNN sequential architecture,
implemented using $\sim$1 million images from
the Labeled Faces in the Wild \cite{[Huang07]}
and YouTube Faces \cite{[Wolf11]} datasets.
Since VGG-Face is trained for classifying faces, we believe that
it can provide effective recognition with the periocular region as well,
given that this region appears in the training images.
Introduced later,
the ResNet networks \cite{[He16]} presented the concept of
residual connections to ease the training of CNNs. By reducing
the number of training parameters, they can be substantially deeper.
The key idea of residual connections is to make available the input of a lower layer
to a higher layer, bypassing intermediate ones.
There are different variants of ResNet networks, depending on its depth.
In this work, we employ ResNet101, having a depth of 347 layers
(including 101 convolutional layers).
In DenseNet networks \cite{[Huang17]}, the residual concept is taken even further
since the feature maps of all preceding layers of a Dense block are
used as inputs of a given layer, and its own feature maps are used as inputs
into all subsequent layers. This encourages feature reuse
throughout the network.
Similarly to ResNet, there are different variants of DenseNet (defined by its depth).
In this paper, we employ Densenet201, having a depth of 709 layers
(including 201 convolutional layers).

In using these networks,
%
%
periocular images are fed into the feature extraction pipeline of each pre-trained CNN \cite{[Nguyen18],[Hernandez18]}.
However, instead of using the vector from the last layer, we employ as feature descriptor
the vector from the intermediate layer identified as the one providing the best performance.
These will be found in the respective experimental sections.
%
%
This approach allows the use of powerful architectures pre-trained with
a large number of images in a related domain, eliminating the need of
designing or re-training a new network for a specific task,
which may be infeasible in case of lack of large-scale databases
in the target domain
(as in the case of periocular recognition with images from different sensors).
The extracted CNN vectors can be simply compared with distance measures.
In our case, we employ the $\chi^2$ distance, which has proven to provide better
results than other measures such as the cosine or Euclidean distances \cite{[Hernandez18]}.

\section{Score Fusion Methods}
\label{sect:fusion-methods}

A biometric verification comparator can be defined as a pattern recognition machine that,
by comparing two (or more) samples of input signals, is designed to recognize two different classes.
The two hypotheses or classes defined for each comparison are \textit{target} hypothesis ($\theta_t$: the compared
biometric data comes from the same individual) and
\textit{non-target} hypothesis ($\theta_{nt}$: the compared data comes from different
individuals). As a result of the comparison,
the biometric system outputs a real number $s$ known as \textit{score}.
The higher the score, the more it supports the target hypothesis,
and vice-versa.
The acceptance or rejection of an individual is based on a decision threshold $\tau$,
and this threshold depends on the priors and decision costs involved in the decision-making process.
However, if we do not know the distributions of target or non-target scores from such comparator or any
threshold, we will not be able to classify the associated biometric samples in general.

Integration at the score level
is the most common approach used in multibiometric systems due to
the ease in accessing and combining the scores ${\bf{s}} = \left( {{s_1}, \ldots ,{s_i}, \ldots ,{s_N}} \right)$ generated by $N$ different comparators \cite{[Fierrez18]}.
Unfortunately, each biometric comparator outputs scores that are in a range that is specific to the comparator,
so score normalization is needed to transform these scores
into a common domain prior to the fusion \cite{[Jain05]}, e.g. $s_i\in [0,1]$ or $s_i\in [-1,1]$, $\forall i \in \left\{ {1, \ldots ,N} \right\}$.
But even if two comparators output scores in the same range,
the same output value (say ${s_i}={s_j}=0.5$ for $i \ne j$)
might not favour the target or non-target hypotheses with
the same strength. The same can be said about the fusion of such scores.
From this viewpoint, outputs are dependent on the comparator, and thus, the
acceptance/rejection decision also depends on the comparator.

These problems can be addressed with the concept of \textit{calibrated} scores.
During calibration, the scores ${\bf{s}} = \left( {{s_1}, \ldots ,{s_i}, \ldots ,{s_N}} \right)$
are mapped to a log-likelihood-ratio (LLR) as
$
    {s^{cal}} \approx \log \left( {\frac{{p\left( {{\bf{s|}}{\theta _t}} \right)}}{{p\left( {{\bf{s|}}{\theta _{nt}}} \right)}}} \right)
$,
where ${s^{cal}}$
represents the calibrated score. Then, a decision can be taken using the Bayes decision rule \cite{[DudaHartStork]}:

\begin{equation}
\text{For a given } \bf{s} \left\{
\begin{gathered}
\text{decide} \:\: \theta_t : (p \left(  \left. \bf{s} \right|
\theta_t \right)/ p \left( \left. \bf{s} \right| \theta_{nt}
\right))
> \tau_B \\
\text{decide} \:\: \theta_{nt} :
(p \left(  \left. \bf{s} \right| \theta_t \right)/ p \left( \left.
\bf{s} \right| \theta_{nt} \right))
< \tau_B \hfill\\
\end{gathered} \right.
\label{eq:bayesDecisionLR}
\end{equation}

The parameter $\tau_B$ is known as the \emph{Bayes threshold}, and
its value depends on the prior probabilities of the hypotheses $p
\left( \theta_t \right)$ and $p \left( \theta_{nt} \right)$ and on
the decision costs.
This form of output
is \textit{comparator-independent}
since this log-likelihood-ratio output can theoretically be used to make
optimal (Bayes) decisions for any given target prior and any costs associated with making erroneous decisions \cite{[DudaHartStork]}.
%
Therefore, the calibration process gives \textit{meaning} to ${s^{cal}}$.
In a Bayesian context,
a calibrated score ${s^{cal}}$ can be interpreted as a degree of support
to any of the hypotheses. If ${s^{cal}}>0$, then
the support to ${\theta _t}$ is also higher, and vice-versa.
Also, the meaning of a log-likelihood ratio is the same across different
biometric comparators, allowing to compare them in the same probabilistic range.
This calibration transformation then solves the two previously commented problems.
First, it maps scores from biometric comparators to a common domain. Second, it
allows the interpretation of biometric scores as a degree of support.

A number of strategies can be used to train a calibration transformation \cite{brummer06}.
Among them, logistic regression has been successfully used for biometric applications \cite{[pigeon00],[brummer07fusion],[Gonzalez-Rodriguez07a],[Ferrer08],[Alonso10]}.
With this method, the scores of
multiple comparators are fused together, primarily to improve the discriminating ability,
in such a way as to encourage good calibration of the output scores.
Given $N$ biometric comparators which
output the scores ${\bf{s}}_j = (s_{1j}, s_{2j}, ... s_{Nj})$ for an input trial
$j$, a linear fusion of these scores is:

\begin{equation}
f_j = a_0 + a_1 \cdot s_{1j} + a_2 \cdot s_{2j} + ... + a_N \cdot s_{Nj}
\label{eq:LLRfusion}
\end{equation}

When the weights $\{a_0,...,a_N\}$
are trained via logistic regression, the fused score $f_j$ 
is a
well-calibrated log-likelihood-ratio
\cite{brummer06,[brummer07fusion]}.
Let $[s_{ij}]$ be an $N \times N_T$ matrix of training scores built from $N$
biometric comparators and $N_T$ target trials, and let $[r_{ij}]$ be an
$N \times N_{NT}$ matrix of training scores built from the same $N$ biometric
comparators with $N_{NT}$ non-target trials. We use a logistic
regression objective \citep{[pigeon00],[brummer07fusion]} that is
normalized with respect to the proportion of target and non-target
trials ($N_T$ and $N_{NT}$, respectively), and weighted with respect
to a given prior probability $P = P\left( {{\rm{target}}} \right)$.
The objective is stated in terms of a \emph{cost} $C$, which must be
\emph{minimized}:

\begin{equation}
 C  = \frac{P}{N_T}\sum\limits_{j = 1}^{N_T} {\log
\left( {1 + e^{ - f_j  - \rm{logit} \emph{P}} } \right) +} {\frac{{1
- P}}{N_{NT}}} \sum\limits_{j = 1}^{N_{NT}} {\log \left( {1 + e^{ -
g_j - \rm{logit} \emph{P}} } \right)} \end{equation}

\noindent where the fused target and non-target scores are
respectively

\begin{equation}
\begin{array}{l}
f_j  = a_0  + \sum\limits_{i = 1}^N {a_i s_{ij} } \\ \\
g_j  = a_0 + \sum\limits_{i = 1}^N {a_i r_{ij} }
\end{array}
\end{equation}

\noindent and where
$
\rm{logit} \emph{P} = \log \left( {\frac{\emph{P}}{{1 - \emph{P}}}}
\right)
$.

It can be demonstrated that minimizing the objective $C$ with
respect to $\{a_0,...,a_N\}$ 
results in a good calibration of the
fused scores \citep{brummer06,[brummer07fusion]}. In practice,
changing the value of $P$ has a small effect. The
default of $0.5$ is a good choice for a general application and it
will be used in this work. The optimization objective $C$ is convex
and therefore has a unique global minimum.
%

%
Another advantage of this method is that when fusing scores from different comparators,
the most reliable comparator will implicitly be given a dominant role in the fusion
(via the trained weights $\{a_0,...,a_N\}$).
In other standard fusion methods,
such as the average of scores \cite{[Jain05]},
all comparators are given the same weight in the fusion,
regardless of its individual accuracy.
%
%
It is also straightforward to show that if
M calibrated scores $\{s^{cal}_{1}
,s^{cal}_{2} , \ldots ,s^{cal}_{M}\}$ come from statistically
independent sources (such as multiple biometric comparators),
its sum $s^{cal}_{1} + s^{cal}_{2} + \ldots +  s^{cal}_{M}$ also yields a log-likelihood ratio
\cite{[DudaHartStork]}.
%
%
The latter allows to calibrate the scores ${s_i}$ of each available biometric comparator separately
(by using $N$=1 in Equation~\ref{eq:LLRfusion}),
and simply sum the calibrated scores ${s^{cal}_i}$ of each comparator in order to obtain a new
calibrated score, as shown in Figure~\ref{fig:fusion_model}.
In this paper, the two possibilities are evaluated, i.e. calibrating the scores of all comparators together vs calibrating them separately and then summing them up.
%
%
%
In order to perform logistic regression calibration, the
freely available Bosaris toolkit for Matlab has been used\footnote{https://sites.google.com/site/bosaristoolkit/}.
For further details of this fusion method,
the reader is referred to \cite{[Alonso10]} and the references therein.

The probabilistic fusion method described above is compared in the present work with three strategies. Since each biometric comparator usually outputs scores that are in a range that is specific to the system, the scores of each comparator are normalized prior to the fusion using z-score normalization \cite{[Jain05]}. The three strategies are:

\begin{itemize}
  \item \textbf{Average}. With this simple rule, the scores of the different comparators are simply averaged.
        Motivated by their simplicity, simple fusion rules have been used in biometric authentication
        with very good results \cite{[Bigun97],[Kittler98]}. They have the advantage of not needing training,
        sometimes surpassing other complex fusion approaches \cite{[Fierrez05d]}.

  \item \textbf{SVM}. Here, a Support Vector Machine (SVM) is trained to provide a binary classification
        given a set of scores from different biometric comparators \cite{[Vapnik95]}.
        The SVM algorithm searches for an optimal hyperplane
        that separates the data into two classes.
        SVM is a popular approach employed in multibiometrics \cite{[Gutschoven00svmfusion]}, which has shown to outperform other trained approaches \cite{[Fierrez18]}.
        In this work, we evaluate Linear, RBF, and Polynomial (order 3) kernels.
        Instead of using the binary predicted class label,
        we use the 
        signed distance to the decision boundary as the output score of the fusion.
        This allows the presentation of DET curves and associated EER and FRR measures.
  \item \textbf{Random Forest}. Another method employed for the fusion of scores
        from multiple biometric comparators is the Random Forest (RF) algorithm \cite{[Ma05rffusion]}.
        An extension of the standard classification tree algorithm, the RF algorithm is an ensemble
        method where the results of many decision trees are combined \cite{[Breiman01rf]}.
        This helps to reduce overfitting and to improve generalization capabilities.
        The trees in the ensemble are grown by using bootstrap samples of the data.
        In this work, we evaluate ensembles with 25, 150, and 600 decision trees.
        Instead of using the binary predicted class label,
        we use the weighted average of the class posterior probabilities
        over the trees that support the predicted class, so we can present DET curves
        and associated measures.
\end{itemize}

\begin{figure*}[htb]
\centering
    \begin{subfigure}[b]{0.45\textwidth}
        \includegraphics[width=\textwidth]{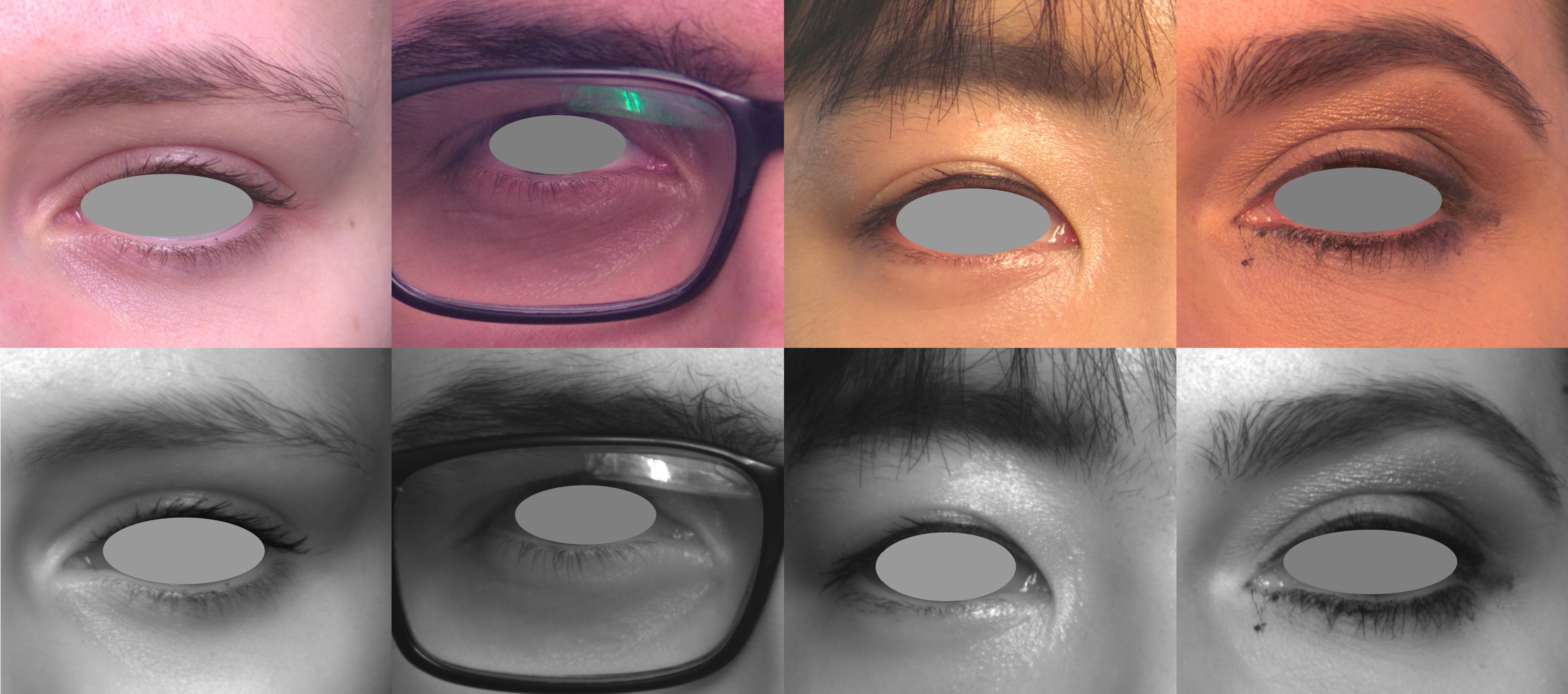}
        \caption{Cross-Eyed database (top row: visible images, bottom: near-infrared)}
    \end{subfigure}
    \begin{subfigure}[b]{0.45\textwidth}
        \includegraphics[width=\textwidth]{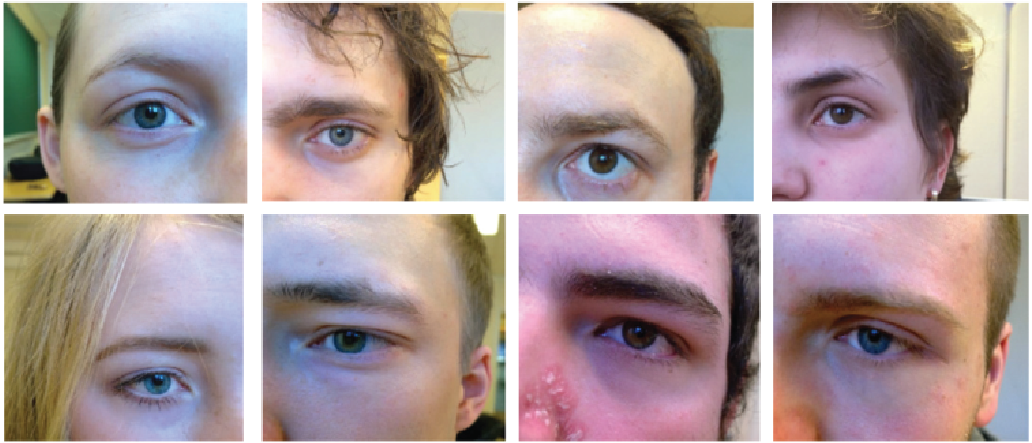}
        \caption{VSSIRIS database (top row: Apple iPhone 5S, bottom: Nokia Lumia 1020; images taken from \cite{[Raja14b]})}
    \end{subfigure}
\caption{Sample periocular images.} \label{fig:db-samples}
\end{figure*}

\section{Cross-Spectral (NIR-VIS) Periocular Recognition}
\label{sect:cross-eyed}

\subsection{Database and Protocol}
\label{sect:cross-eyed-dbprotocol}

In the cross-spectral recognition experiments of this section,
we employ the Reading
Cross-Spectral Iris/Periocular Dataset 
used as the
benchmark dataset for the 1$^{st}$ Cross-Spectral Iris/Periocular
Competition (Cross-Eyed 2016) \cite{[sequeira16crosseyed]}.
The dataset contains both visible (VIS) and near-infrared (NIR)
images captured with a custom dual spectrum imaging sensor which
acquires images in both spectra synchronously.
Periocular images are of size $800\times900$ (height$\times$width)
from 120 subjects, with 8 images of both eyes captured in both
spectra, totalling 3840 images.
Images are captured at a distance of 1.5 m, in an uncontrolled indoor
environment, containing large variations in ethnicity, eye colour,
and illumination reflections.
Some examples are shown in Figure~\ref{fig:db-samples} (top). To
avoid usage of iris information by periocular methods during the
Cross-Eyed competition, periocular images were distributed with a
mask on the eye region, as discussed in \cite{[Park11]}.
A new edition of the competition was held in 2017.
The 120 subjects of the Cross-Eyed 2016 database were provided
as the training set, and an additional set of
55 subjects were sequestered as the test set in the 2017 edition,
but the latter was never released \cite{[sequeira17crosseyed]}.

\begin{table}[htb]
\scriptsize
\begin{center}
\begin{tabular}{cccc}

 \multicolumn{4}{c}{} \\

 \multicolumn{4}{c}{\textbf{Cross-Eyed database}} \\ \hline

\multicolumn{2}{c}{\textbf{Comparison type}} &
 \multicolumn{1}{c}{\textbf{Training}} &
 \multicolumn{1}{c}{\textbf{Test}}\\

 \multicolumn{2}{c}{} &
 \multicolumn{1}{c}{(30 subjects)} &
 \multicolumn{1}{c}{(90 subjects)}\\ \hline \hline

Same- & Genuine  & 30 $\times$ 2E $\times$ (7+6+...+1) = 1,680 &
90 $\times$ 2E $\times$ (7+6+...+1) = 5,040  \\

Sensor & Impostor  & 30 $\times$ 29 $\times$ (4L + 4R) =
6,960 &
90 $\times$ 89 $\times$ (2L + 2R) = 32,040  \\
\hline \hline

Cross- & Genuine  & 30 $\times$ 2E $\times$ 8L $\times$ 8R = 3,840 &
90 $\times$ 2E $\times$ 8L $\times$ 8R =11,520 \\

 Spectral & Impostor  & 30 $\times$ 29 $\times$ (4L+4R) $\times$ 2S = 13,920 &
 90 $\times$ 89 $\times$ (2L+2R) $\times$ 2S = 64,080 \\

\hline

\end{tabular}

\end{center}
\caption{Cross-Eyed database: Experimental protocol. E=Eyes, L=Left eye, R=Right eye, S=Sensors.}
\label{tab:exp-protocol-crosseyed}
\end{table}
\normalsize

\begin{table}[htb]
\scriptsize
\begin{center}
\begin{tabular}{cccc}

\multicolumn{4}{c}{} \\


\multicolumn{1}{c}{\textbf{comparator}} &
\multicolumn{1}{c}{\textbf{Cross-Eyed}} &
\multicolumn{1}{c}{\textbf{VSSIRIS}} &
\multicolumn{1}{c}{\textbf{data}}\\ \hline

\multicolumn{1}{c}{SAFE} & 6$\times$3$\times$9=162  & 6$\times$3$\times$9=162  & complex  \\
\hline

\multicolumn{1}{c}{GABOR} & 48$\times$30=1440  & 56$\times$30=1680  & real  \\
\hline

\multicolumn{1}{c}{SIFT (AR original)} & circa 243200  & circa 384000  & real  \\
\hline

\multicolumn{1}{c}{SIFT (FR original)} & circa 325504  & circa 489472 & real  \\
\hline

\multicolumn{1}{c}{SIFT (FR 224$\times$224)} & circa 11776 & circa 16512  & real  \\
\hline

\multicolumn{1}{c}{LBP, HOG} & 48$\times$8=384  & 56$\times$8=448  & real  \\
\hline

\multicolumn{1}{c}{NTNU} & 9472  & 9472  & integer  \\
\hline

\multicolumn{1}{c}{VGG-Face} & 100352  & 100352  & real  \\
\hline

\multicolumn{1}{c}{Resnet101} & 50176  & 100352  & real  \\
\hline

\multicolumn{1}{c}{Densenet201} & 6272  & 43904  & real  \\
\hline

 \multicolumn{4}{c}{} \\

%
%
%
%
%
%
%
%
%

\end{tabular}

\end{center}
\caption{Size of the feature vector per comparator and per database. AR=annular ROI. FR=Full ROI. 'Original' refers to the original size of the input image (Cross-Eyed: $613\times701$, VSSIRIS: 871$\times$871).}
\label{tab:vector-size}
\end{table}
\normalsize

\begin{table}[htb]
\scriptsize
\begin{center}
\begin{tabular}{|c|cc|cc|}

 \multicolumn{5}{c}{} \\

 \multicolumn{1}{c}{} & \multicolumn{2}{c}{\textbf{Cross-Eyed database}} & \multicolumn{2}{c}{\textbf{VSSIRIS database}} \\ \cline{2-5}

\multicolumn{1}{c}{\textbf{}} &
 \multicolumn{1}{|c}{\textbf{Extraction}} &
  \multicolumn{1}{c|}{\textbf{Comparison}} &
 \multicolumn{1}{c}{\textbf{Extraction}} &
  \multicolumn{1}{c|}{\textbf{Comparison}}   \\

\multicolumn{1}{c}{\textbf{}} &
\multicolumn{1}{|c}{\textbf{Time}} &
 \multicolumn{1}{c|}{\textbf{Time}} &
 \multicolumn{1}{c}{\textbf{Time}} &
  \multicolumn{1}{c|}{\textbf{Time}}

 \\ \hline

SAFE   & 2.98 sec & 0.2 ms & 11.86 sec & $<$0.1 ms \\ \hline

GABOR   & 0.49 sec & 0.3 ms  & 0.53 sec & 0.3 ms  \\
\hline

SIFT (AR original)   & 0.94 sec & 0.58 sec & 1.5 sec & 1.1 sec \\
\hline

SIFT (FR original)  & 0.94 sec & 0.94 sec & 1.5 sec & 1.7 sec \\
\hline

SIFT (FR 224$\times$224) & 0.05 sec & 1.6 ms & 0.07 sec & 3 ms \\
\hline

LBP   & 0.16 sec & $<$0.1 ms & 0.17 sec & $<$0.1 ms \\
\hline

HOG   & 0.01 sec & $<$0.1 ms & 0.13 sec & $<$0.1 ms \\
\hline

NTNU   & 0.6 sec & 0.7 ms & 0.56 sec & 0.7 ms \\
\hline

VGG-Face   & 0.51 sec & 1.65 ms & 0.52 sec & 1.43 ms \\
\hline

Resnet101  & 0.27 sec & 0.35 ms & 0.48 sec & 0.65 ms \\
\hline

Densenet201  & 0.25 sec & $<$0.1 ms & 0.39 sec & 0.42 ms \\
\hline

\multicolumn{5}{c}{} \\

\end{tabular}
\end{center}
\caption{Feature computation times for each database. AR=annular ROI. FR=Full ROI. 'Original' refers to the original size of the input image (Cross-Eyed: 613$\times$701, VSSIRIS: 871$\times$871).}
\label{tab:file-size-time}
\end{table}
\normalsize

Prior to the competition, a \emph{training} set of images
from 30 subjects was distributed.
The \emph{test} set consisted of images from 80
subjects, sequestered by the organizers and distributed
after the competition.
Images from 10 additional subjects were also released after the
competition that were not present in the test set.
Here, we will employ the same 30 subjects of the training set to
tune our algorithms and the remaining 90 subjects for testing
purposes.
All images have an annotation mask of the eye region.
%
The mass centre of the mask is set as the reference point (centre)
of the eye. Images are then rotated w.r.t. the axis that crosses
the two mask corners and resized via bicubic interpolation to
have the same corner-to-corner distance (set to 318
pixels, the average value of the training set). Then, images
are aligned by extracting a region of 613$\times$701 around the eye.
This size is set empirically to ensure that all available images
have sufficient margin to the four sides of the eye centre.
Eyes in the Cross-Eyed database are slightly displaced in the vertical
direction, so the eye is not centred in the aligned images but with
a vertical offset of 56 pixels (see Figure~\ref{fig:img-ROI}, top).
Images are further processed by Contrast Limited Adaptive Histogram
Equalization (CLAHE) \cite{[Zuiderveld94clahe]}, which is the
preprocessing choice with ocular images \cite{[Rathgeb10]}, and then
sent to feature extraction. 

We carry out verification experiments, with each eye considered a
different user. 
%
We compare images both from the same device (\emph{same-sensor}) and
from different devices (\emph{cross-spectral}). Genuine trials are
obtained by comparing each image of an eye to the remaining images
of the same eye. 
In \emph{same-sensor} comparisons, to avoid symmetric comparisons, the first image of an eye is compared to the second to eight images; the second image is compared to the third to eight images, and so on, leading to (7+6+...+1) genuine scores per eye.
This procedure is repeated for the two eyes of all subjects.
This results in 30 subjects $\times$ 2 eyes $\times$ (7+6+...+1) and 90 $\times$ 2 $\times$ (7+6+...+1) genuine scores with the training and test set, respectively.
In \emph{cross-spectral} comparisons, the eight images of an eye in one spectrum are compared against the eight images in the other spectrum, leading to 8 $\times$ 8 genuine scores per eye.
This results in 30 subjects $\times$ 2 eyes $\times$ 8 $\times$ 8 and 90 $\times$ 2 $\times$ 8 $\times$ 8 genuine scores with the training and test set, respectively.
Impostor trials are done by comparing the 1$^{st}$ image of an
eye to the 2$^{nd}$ image of the remaining eyes.
In \emph{same-sensor} comparisons, given a subject of the test set, his/her 1$^{st}$ image of both eyes is compared against the 2$^{nd}$ image of both eyes from the remaining 89 subjects. This results in 89 $\times$ 4 test impostor scores per subject and 90 $\times$ 89 $\times$ 4 impostor scores in total.
In \emph{cross-spectral} comparisons, the number of impostor scores is doubled by comparing the 1$^{st}$ image in VIS against the 2$^{nd}$ image in NIR, and the 1$^{st}$ image in NIR against the 2$^{nd}$ image in VIS. This results in 90 $\times$ 89 $\times$ 4 $\times$ 2 test impostor scores in total.
To increase the
number of available training scores, we carry out an additional
comparison to the 3$^{rd}$ image of the remaining eyes only with the
training set, effectively duplicating the number of impostor scores per subject.
Since the training set contains 30 subjects, this results in 29 $\times$ 4 $\times$ 2 (\emph{same-sensor}) and 29 $\times$ 4 $\times$ 2 $\times$ 2 (\emph{cross-spectral}) training impostor scores per subject. By multiplying these amounts by 30, we obtain the total amount of impostor scores with the training set.
The experimental protocol is summarized in
Table~\ref{tab:exp-protocol-crosseyed}.

The periocular comparators employed have some parameters
which are set as follows.
It should be highlighted
that these parameters are computed
in proportion to the size of the image, without any other training.
If the image size changed, they would adapt dynamically
so that the comparators would always be capturing their features in the same relative areas of the image.
The only input needed is the position of the eye corners,
which were also used to normalize and crop the image to a constant size, as described above.
Regarding the SAFE comparator, the annular band of the first
circular ring starts at a radius of $R$=79 pixels (determined
empirically as 1/4 of the eye corner-to-corner distance), and the
band of the last ring ends at the bottom boundary of the image. This
results in a ROI of $501\times501$ pixels around the eye centre
(as shown in Figure~\ref{fig:img-ROI}, third column).
The grid employed with GABOR, LBP and HOG comparators has 7$\times$8=56 non-overlapping
blocks. Based on the size of the input image, each block has 88$\times$88 pixels.
The 8 central blocks are not considered since they are equal for all users due to the eye region mask, so
features are extracted only from 48 blocks.
The GABOR comparator employs filter wavelengths spanning from 44 to 6
pixels, which are set proportional to the block size as 88/2=44 to
88/16$\approx$6. 
The VGG-Face, Resnet101, and Densenet201 comparators employ an input image size of 224$\times$224,
so images are resized to match these dimensions.
Regarding the SIFT comparator, our baseline configuration entails the use of the same annular ROI than the SAFE comparator \cite{[Alonso17b_eusipco_busch]} for key-point extraction. However, for comparison purposes with the other systems, we also evaluate the use of the entire input image (except the 8 central blocks). This is done both at the original size of the image (613$\times$701) and at the input size of the CNNs (224$\times$224).
Table~\ref{tab:vector-size} (second column)
indicates the size of the feature vector
for a given periocular image with the different comparators employed.
Obviously, the SIFT descriptor is dependant on the size of the ROI and the image. With the full ROI, the average number of key-points per image is 2543 (of which a vector of 128 elements is computed, resulting in 2543$\times$128=325504 real values per image). The annular ROI produces a slightly smaller amount (1900 key-points, or 243200 values) and if the image is reduced to 224$\times$224, the amount is substantially less (only 92 key-points, or 11776 values).
Experiments have been done in a Dell Latitude E7240 laptop with an
i7-4600 (2.1 GHz) processor, 16 Gb DDR3 RAM, and a built-in Intel HD
Graphics 4400 card. The OS is Microsoft Windows 8.1 Professional,
and the comparators are implemented in Matlab 
x64, with the
exception of SIFT that is implemented in C++ and invoked from Matlab
via MEX files.
The VGG-Face model is from Caffee, which has been imported to Matlab
with the \texttt{importCaffeNetwork} function.
The Resnet101 and Densenet201 models are from the pre-trained models available
in Matlab r2019a.
In line with the Cross-Eyed competition, we also provide the extraction and
comparison time of each method 
%
(Table~\ref{tab:file-size-time}, second and third columns).
Here, it can be also appreciated the variation of the SIFT versions depending on the image or ROI size.

\begin{figure}[htb]
\centering
        \includegraphics[width=0.95\textwidth]{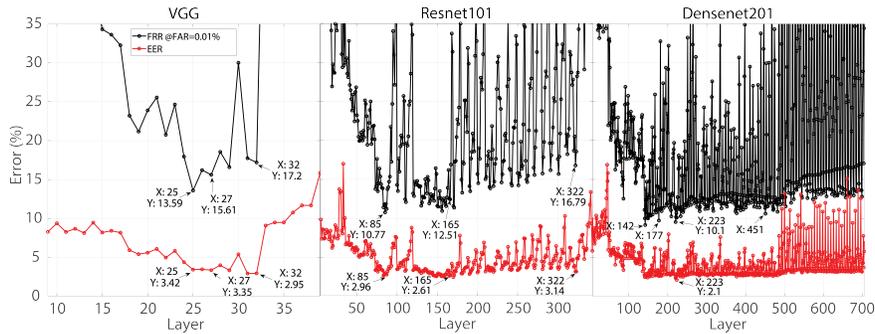}
\caption{Cross-Eyed database: Cross-spectral accuracy (VIS-NIR) of different CNN layers.}
\label{fig:crosseyed-results-cnns}
\end{figure}


\begin{figure}[htb]
\centering
        \includegraphics[width=0.45\textwidth]{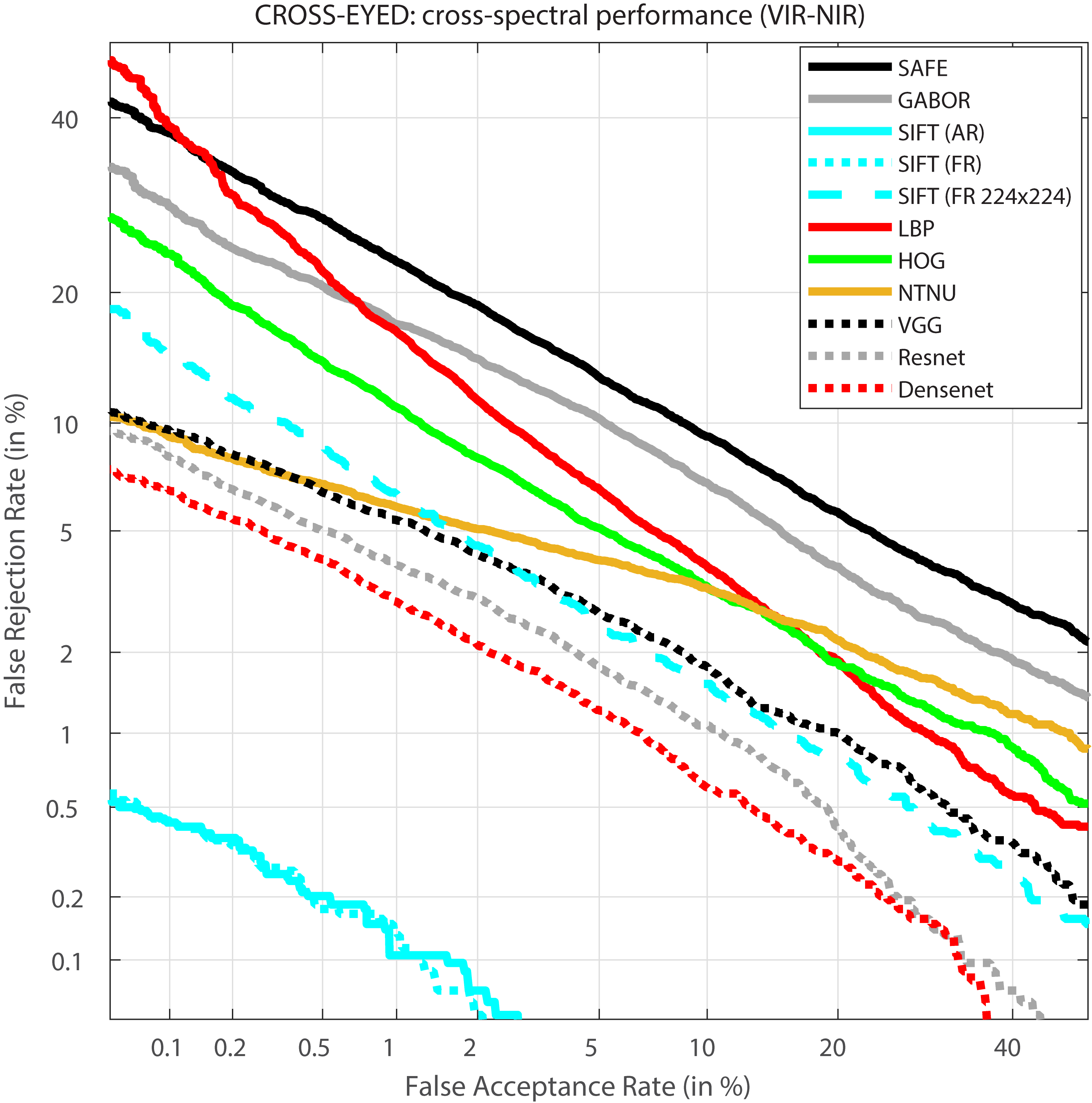}
\caption{Cross-Eyed database, test set: Verification results of the individual comparators. Best seen in colour.}
\label{fig:crosseyed-results1}
\end{figure}

\subsection{Results: Finding the Optimum Layer of the Convolutional Neural Networks}
\label{subsec:best-CNN-layer-crosseyed}

Normalized periocular images are fed into the feature extraction of each pre-trained CNN.
We investigate the representation capability of each layer by reporting the corresponding
cross-spectral accuracy using features from each layer. The recognition accuracy of each network (EER and FRR @ FAR=0.01\%) is given in Figure~\ref{fig:crosseyed-results-cnns}.
It is worth noting that the best performance is obtained in some intermediate layer for all
CNNs, in line with previous studies using ocular modalities \cite{[Nguyen18],[Hernandez18]}.
In selecting the best layer, we prioritize the FRR @ FAR=0.01\%, since this was the metric
employed to rank submissions to the Cross-Eyed competition, although we seek a balance with
the EER as well.
We have also searched for layers that give optimum performance both with the Cross-Eyed and
the VSSIRIS databases simultaneously if possible
(results with the latter are given in Figure~\ref{fig:vssiris-results-cnns}).

A good performance with VGG-Face is obtained with layer 25, 
which is a max pooling layer with $14\times14\times512=100352$ elements.
Layer 27 also provides good performance.
This is a ReLu layer of the same size as layer 25,
but since it has many elements set to 0 due to the ReLu operation,
we prefer to choose layer 25.
VGG-Face is a serial network, with layers arranged one after the other.
On the other hand, ResNet101 and Densenet201 are acyclic networks,
in which layers have inputs from multiple layers and outputs to
multiple layers. This more intricate architecture may thus explain
the oscillations observed between layers.
With ResNet101, a good performance is obtained with layer 165.
This is a convolutional layer with $14\times14\times256=50176$ elements,
and it will be the layer employed with Cross-Eyed.
With VSSIRIS, better performance is obtained with layer 323,
which is not the case with Cross-Eyed.
This is
a ReLu layer with $7\times7\times2018=100352$ elements.
We choose this layer with VSSIRIS instead since it provides better EER
than other layers as well.
Regarding DenseNet201, good performance with Cross-Eyed (which minimizes
both the EER and FRR) is obtained with layer 223.
This is a convolutional layer with only $14\times14\times32=6272$ elements.
Other layers (e.g. 142 or 177) also give a good FRR, but the EER is
not as good as with layer 223.
With VSSIRIS, better performance is given by layer 480 instead,
which is an average pooling layer with $7\times7\times896=43904$ elements.

\begin{table}[htb]
\scriptsize
\begin{center}
\begin{tabular}{ccccccccc}

\multicolumn{9}{c}{\textbf{}} \\

 \multicolumn{2}{c}{} &  \multicolumn{3}{c}{\textbf{Equal Error Rate (EER)}}
 & \multicolumn{1}{c}{} &  \multicolumn{3}{c}{\textbf{FRR @
 FAR=0.01\%}}\\ \cline{3-5} \cline{7-9}

 \multicolumn{1}{c}{\textbf{}} & \multicolumn{1}{c}{\textbf{}} &
\multicolumn{2}{c}{\textbf{Same sensor}}
& \multicolumn{1}{c}{\textbf{}}
& \multicolumn{1}{c}{\textbf{}} & \multicolumn{2}{c}{\textbf{Same sensor}}  & \multicolumn{1}{c}{\textbf{}}\\
 \cline{3-4} \cline{7-8}

\multicolumn{1}{c}{\textbf{comparator}} & \multicolumn{1}{c}{\textbf{}} &
 \multicolumn{1}{c}{\textbf{NIR}} &
 \multicolumn{1}{c}{\textbf{VIS}} &
 \multicolumn{1}{c}{\textbf{cross-spectral}} & &
 \multicolumn{1}{c}{\textbf{NIR}} &
 \multicolumn{1}{c}{\textbf{VIS}} &
 \multicolumn{1}{c}{\textbf{cross-spectral}}
 \\ \cline{1-1} \cline{3-5} \cline{7-9}

SAFE & & 5.85\% & 5.67\% & 9.47\% (+67\%) &  & 22.4\% & 24.23\% & 50.38\% (+124.9\%)
 \\ \cline{1-1} \cline{3-5} \cline{7-9}

GABOR & & 5.48\% & 5.34\% & 7.94\% (+48.7\%) &  & 26.25\% & 23.68\% & 43.3\% (+82.9\%)
 \\ \cline{1-1} \cline{3-5} \cline{7-9}

SIFT (AR 613$\times$701) & & 0.02\% & 0\% & 0.28\% ($>$1300\%) &  & 0.02\% & 0\% & 0.88\% ($>$4300\%)
 \\ \cline{1-1} \cline{3-5} \cline{7-9}

SIFT (FR 613$\times$701) & & 0.02\% & 0\% & 0.27\% ($>1250$\%) &  & 0.02\% & 0\% & 0.9\% ($>4400$\%)
 \\ \cline{1-1} \cline{3-5} \cline{7-9}

SIFT (FR 224$\times$224) & & 1.11\% & 0.86\% & 3.36\% ($+290$\%) &  & 5.83\% & 2.98\% & 29.7\% ($+897$\%)
 \\ \cline{1-1} \cline{3-5} \cline{7-9}



LBP & & 3.03\% & 3.27\% & 5.84\% (+92.7\%) & & 10.97\% & 12.86\% & 63.79\% (+481.5\%)
 \\ \cline{1-1} \cline{3-5} \cline{7-9}

HOG & & 3.84\% & 4.19\% & 5.06\% (+31.8\%) &   & 11.76\% & 14.93\% & 34.36\% (+192.2\%)
 \\ \cline{1-1} \cline{3-5} \cline{7-9}

NTNU & & 2.83\% & 2.45\% & 4.22\% (+72.2\%) & & 3.93\% & 3.57\% & 13.8\% (+286.6\%)
 \\ \cline{1-1} \cline{3-5} \cline{7-9}

VGG-Face & & 2.36\% & 2.53\% & 3.42\% (+44.9\%) &  & 8.48\% & 8.68\% & 13.59\% (+60.3\%)
 \\ \cline{1-1} \cline{3-5} \cline{7-9}

Resnet101 & & 1.52\% & 1.6\% & 2.61\% (+71.7\%) &  & 5.51\% & 5.01\% & 12.51\% (+149.7\%)
 \\ \cline{1-1} \cline{3-5} \cline{7-9}

Densenet201 & & 1.37\% & 1.54\% & 2.09\% (+52.6\%) &  & 5.69\% & 5.18\% & 10.09\% (+94.8\%)
 \\ \cline{1-1} \cline{3-5} \cline{7-9}

\end{tabular}

\end{center}
\caption{Cross-Eyed database, test set: Verification results of the individual
comparators. The relative variation of cross-spectral performance with
respect to the best same-sensor performance is given in brackets (for the SIFT rows with VIS=0\%, the result is calculated w.r.t. the NIR performance 
to avoid division by zero). AR=annular ROI. FR=Full ROI.}
\label{tab:crosseyed-results}
\end{table}
\normalsize

\subsection{Results: Individual Comparators}

We now report the performance of all periocular comparators in Table~\ref{tab:crosseyed-results}.
Besides the EER, we also report
the FRR at FAR=0.01\%. The latter
was the metric used to rank submissions to the Cross-Eyed
competition.
We report two types of results: $i$) \emph{same-sensor} comparisons; and
$ii$) \emph{cross-spectral} comparisons. 
%
In Figure~\ref{fig:crosseyed-results1} we also give the DET curves of the
cross-spectral experiments. 

From 
Table~\ref{tab:crosseyed-results}, it can be seen that given a comparator, the NIR and VIS performances (same-sensor) are relatively equal. For example, the EER of SAFE is 5.85\% (NIR) and 5.67\% (VIS), so if the two images are in the same spectrum, there is no significant advantage in operating in NIR or VIS.
This happens with all comparators, both in the EER and the FRR (with just a few exceptions), which is very interesting because they are based on different image features.
%
%
In previous studies, the periocular modality usually performed better with
VIS data \cite{[Alonso15a],[Hollingsworth12],[Woodard10]}, so it is
generally accepted this modality is most suited to VIS
imagery \cite{[Alonso16]}.
On the contrary, some other works show opposite results
\cite{[Sharma14]}.
However, in the mentioned studies,
the images employed are
of smaller size, ranging from 100$\times$160 to 640$\times$480,
while the images employed in this paper are of 613$\times$701 pixels.
%
Also, they evaluate three different periocular comparators at most.
In the present paper,
the use of bigger images 
may be the reason for a comparable performance between NIR and VIS images.

Regarding cross-spectral experiments, we observe a significant
worsening in performance w.r.t. same-sensor comparisons, although not
all comparators are affected in the same way.
%
%
%
HOG, NTNU and especially LBP are substantially affected in high security mode (i.e. low FAR),
as can be appreciated in the 
right part of Table~\ref{tab:crosseyed-results}.
The relative FRR increase @ FAR=0.01\% for these comparators
is in the range of 200\% to nearly 500\%.
%
But the comparator that is most affected is SIFT.
%
%
%
%
%
%
Even if its cross-spectral performance is the best among all comparators,
it is about one or two orders of magnitude worse than its same-sensor performance (meaning a thousand per cent worse or more).
This is despite the use of
a descriptor with a bigger size (see Table~\ref{tab:vector-size}).
%
%
SIFT extracts features from a discrete set of local key-points, but it might be that the position of detected key-points is not the same in each spectrum.
With the other comparators, on the other hand, the image is divided into annular or square regions (Figure~\ref{fig:img-ROI}), and features are extracted from each region, ensuring a consistent extraction between both spectra.
%
%

Concerning the individual performance of each comparator, SIFT exhibits
very low error rates at the original image size, but this comparator is computationally heavy
both in processing times and template size.
In this paper,
we use the SIFT detector with the same parametrization employed in
\cite{[Alonso09]} for iris images of size 640$\times$480.
In the work \cite{[Alonso09]},
the iris region represented $\sim$1/8 of the image only, leading to
some hundreds of key-points per image.
%
%
However, images of the Cross-Eyed database are of 613$\times$701 pixels,
and the periocular ROI occupies a considerably bigger area than the iris region,
leading to an average of $\sim$1900 key-points per image (annular ROI) or $\sim$2543 (full ROI).
%
To match two images,
it is needed to compare each key-point of one image
against all key-points of the other image to find a pair match.
This increases the computation time exponentially when the number of
key-points per image increases,
which is one of the drawbacks of key-point based comparators \cite{[Alonso16]}.
%
%
%
%
%
%
The other comparators employed have templates of fixed size, thus
comparison is made very efficiently using distance measures involving
a number of fixed calculations.
%
%
It can be also seen that the use of annular (AR) or full ROI (FR) does not produce a significant difference with SIFT. 
This suggests that the annular ROI is sufficient, and the key-points of the corner areas incorporated with the full image (Figure~\ref{fig:img-ROI}) do not contribute to a better performance with the Cross-Eyed database, while the comparison time is increased by 62\% (Table~\ref{tab:file-size-time}).
Therefore, we carry forward the AR configuration of SIFT to the fusion experiments of the next section.
On the other hand, if the size of the input image is reduced to match the CNNs (224$\times$224), the lower amount of detected key-points (only 92 on average) produces that both same-sensor and cross-spectral performance degrades one of two orders of magnitude. When this happens, SIFT becomes worse than e.g. DenseNet201 or ResNet 101, and comparable to VGG-Face in some DET regions (Figure~\ref{fig:crosseyed-results1}).
This can also serve as indication of the strength of the CNNs, which match SIFT's performance if the image size of the latter is reduced to be equal, and they also rank ahead of the other comparators while using a smaller input image size.

In general, there is an inverse proportion between the error rates
and the template size. 
%
The comparators with the best performance (SIFT, NTNU and the three CNNs)
are also the ones with the biggest feature vector (see Table~\ref{tab:vector-size}).
It is remarkable the performance of NTNU as well, surpassing the CNNs in some cases, but
with a smaller feature vector.
When it comes to cross-spectral comparisons, however, the CNNs provide better
performance. This is observed especially with the deeper networks (ResNet101 and DenseNet201),
highlighting the capability of these powerful descriptors pre-trained with millions of images.
%
%
In the DET curves of Figure~\ref{fig:crosseyed-results1}, it can be better appreciated
the superiority of the three CNNs for cross-spectral comparisons
w.r.t. the other comparators (apart from SIFT).
It is also remarkable the behaviour of DenseNet201, which provides the second-best result of all comparators,
but with a feature vector much smaller than the other CNNs.
Among the three CNNs used in this paper,
DenseNet201 is the one providing the best performance on the original task for which they were trained (ImageNet), so it could be expected that such superiority is transferred to other tasks as well.
%
It is also worth noting the relatively good cross-spectral EER values
of some \textit{light} comparators such as LBP or HOG.
With a feature vector of only 384 real numbers and an EER of 5-6\%,
they would enable low-security applications where computational resources are limited.

\begin{figure}[htb]
\centering
        \includegraphics[width=0.95\textwidth]{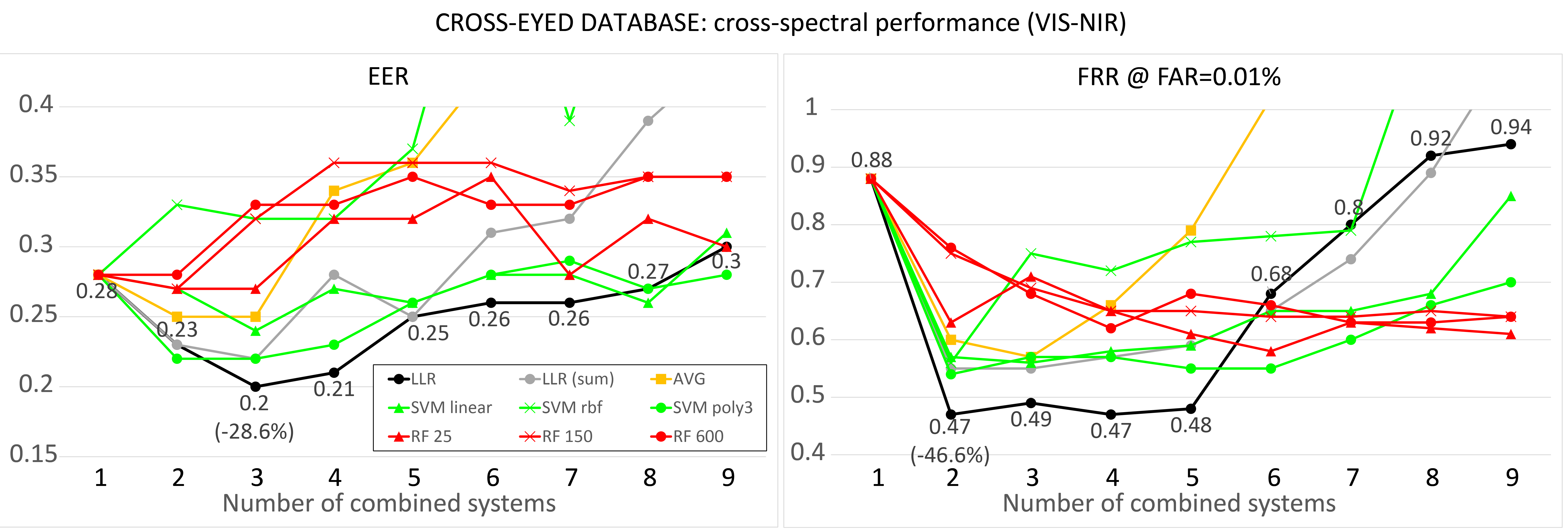}
\caption{Cross-Eyed database, test set: Verification results for an increasing number of fused comparators. Best seen in colour.}
\label{fig:crosseyed-results-fusion}
\end{figure}

\setlength{\tabcolsep}{0pt}

\begin{table}[htb]
\tiny
\begin{center}

\begin{tabular}{|P{0.25cm}|P{0.2cm}|P{0.2cm}P{0.2cm}P{0.2cm}P{0.2cm}P{0.2cm}P{0.2cm}P{0.2cm}P{0.2cm}P{0.2cm}|P{0.7cm}|P{0.7cm}|p{0.2cm}|P{0.2cm}P{0.2cm}P{0.2cm}P{0.2cm}P{0.2cm}P{0.2cm}P{0.2cm}P{0.2cm}P{0.2cm}|P{0.7cm}|P{0.7cm}|p{0.2cm}|P{0.2cm}P{0.2cm}P{0.2cm}P{0.2cm}P{0.2cm}P{0.2cm}P{0.2cm}P{0.2cm}P{0.2cm}|P{0.7cm}|P{0.7cm}|}

\multicolumn{37}{c}{\textbf{CROSS-EYED DATABASE: cross-spectral performance (VIS-NIR)}} \\
\cline{3-13} \cline{15-25} \cline{27-37}

\multicolumn{1}{c}{} & \multicolumn{1}{c}{} & \multicolumn{11}{|c|}{\textbf{LLR FUSION}} & \multicolumn{1}{c}{} & \multicolumn{11}{|c|}{\textbf{AVERAGE FUSION}} & \multicolumn{1}{c}{} & \multicolumn{11}{|c|}{\textbf{SVM LINEAR FUSION}}  \\  \cline{3-13} \cline{15-25} \cline{27-37}




\multicolumn{1}{P{0.12cm}}{\begin{turn}{90}\textbf{\# comparators}
\end{turn}} & \multicolumn{1}{c}{} &
\multicolumn{1}{|P{0.04cm}}{\begin{turn}{90}\textbf{safe}
\end{turn}} &
\multicolumn{1}{P{0.04cm}}{\begin{turn}{90}\textbf{gabor}
\end{turn}} & \multicolumn{1}{P{0.04cm}}{\begin{turn}{90}\textbf{sift}
\end{turn}} & \multicolumn{1}{P{0.04cm}}{\begin{turn}{90}\textbf{lbp}
\end{turn}} &
\multicolumn{1}{P{0.04cm}}{\begin{turn}{90}\textbf{hog}
\end{turn}} & \multicolumn{1}{P{0.04cm}}{\begin{turn}{90}\textbf{ntnu}
\end{turn}} & \multicolumn{1}{P{0.04cm}}{\begin{turn}{90}\textbf{vgg-face}
\end{turn}} & \multicolumn{1}{P{0.04cm}}{\begin{turn}{90}\textbf{resnet101}
\end{turn}} & \multicolumn{1}{P{0.04cm}|}{\begin{turn}{90}\textbf{densenet201}
\end{turn}}
& EER (\%) & FRR (\%) &
\multicolumn{1}{P{0.12cm}}{} & \multicolumn{1}{|P{0.04cm}}{\begin{turn}{90}\textbf{safe}
\end{turn}} &
\multicolumn{1}{P{0.04cm}}{\begin{turn}{90}\textbf{gabor}
\end{turn}} & \multicolumn{1}{P{0.04cm}}{\begin{turn}{90}\textbf{sift}
\end{turn}} & \multicolumn{1}{P{0.04cm}}{\begin{turn}{90}\textbf{lbp}
\end{turn}} &
\multicolumn{1}{P{0.04cm}}{\begin{turn}{90}\textbf{hog}
\end{turn}} & \multicolumn{1}{P{0.04cm}}{\begin{turn}{90}\textbf{ntnu}
\end{turn}} & \multicolumn{1}{P{0.04cm}}{\begin{turn}{90}\textbf{vgg-face}
\end{turn}} & \multicolumn{1}{P{0.04cm}}{\begin{turn}{90}\textbf{resnet101}
\end{turn}} & \multicolumn{1}{P{0.04cm}|}{\begin{turn}{90}\textbf{densenet201}
\end{turn}}
& EER (\%) & FRR (\%) &
\multicolumn{1}{P{0.12cm}}{} & \multicolumn{1}{|P{0.04cm}}{\begin{turn}{90}\textbf{safe}
\end{turn}} &
\multicolumn{1}{P{0.04cm}}{\begin{turn}{90}\textbf{gabor}
\end{turn}} & \multicolumn{1}{P{0.04cm}}{\begin{turn}{90}\textbf{sift}
\end{turn}} & \multicolumn{1}{P{0.04cm}}{\begin{turn}{90}\textbf{lbp}
\end{turn}} &
\multicolumn{1}{P{0.04cm}}{\begin{turn}{90}\textbf{hog}
\end{turn}} & \multicolumn{1}{P{0.04cm}}{\begin{turn}{90}\textbf{ntnu}
\end{turn}} & \multicolumn{1}{P{0.04cm}}{\begin{turn}{90}\textbf{vgg-face}
\end{turn}} & \multicolumn{1}{P{0.04cm}}{\begin{turn}{90}\textbf{resnet101}
\end{turn}} & \multicolumn{1}{P{0.04cm}|}{\begin{turn}{90}\textbf{densenet201}
\end{turn}}
& EER (\%) & FRR (\%) \\
\cline{1-1} \cline{3-13} \cline{15-25} \cline{27-37}

1 &  &  &  & x &  &  &  &  &  &  & 0.28 & 0.88 &  &  &  & x &  &  &  &  &  &  & 0.28 & 0.88  &  &  &  & x &  &  &  &  &  &  & 0.28 & 0.88  \\ \cline{3-13} \cline{15-25} \cline{27-37}
 &  &  &  &  &  &  &  &  &  & x & 2.09 & 10.09 &  &  &  &  &  &  &  &  &  & x & 2.09 & 10.09  &  &  &  &  &  &  &  &  &  & x & 2.09 & 10.09  \\ \cline{3-13} \cline{15-25} \cline{27-37}
 &  &  &  &  &  &  &  &  & x &  & 2.62 & 12.51 &  &  &  &  &  &  &  &  & x &  & 2.62 & 12.51  &  &  &  &  &  &  &  &  & x &  & 2.62 & 12.51  \\ \cline{1-1} \cline{3-13} \cline{15-25} \cline{27-37}

\multicolumn{37}{c}{} \\[-3ex] \cline{1-1} \cline{3-13} \cline{15-25} \cline{27-37}

2 &  &  &  & x &  &  &  &  &  & x & 0.23 & \textbf{0.47} &  &  &  & x &  &  &  &  & x &  & \textbf{0.25} & 0.6  &  &  &  & x &  &  &  & x &  &  & 0.27 & 0.57  \\ \cline{3-13} \cline{15-25} \cline{27-37}
 &  &  &  & x &  &  &  &  & x &  & 0.21 & 0.48 &  &  &  & x &  &  &  &  &  & x & \textbf{0.25} & 0.62  &  &  &  & x &  &  &  &  &  & x & 0.24 & 0.59  \\ \cline{3-13} \cline{15-25} \cline{27-37}
 &  & x &  & x &  &  &  &  &  &  & 0.26 & 0.52 &  &  &  & x &  &  &  & x &  &  & 0.33 & 0.66  &  &  &  & x &  &  &  &  & x &  & 0.23 & 0.61  \\ \cline{1-1} \cline{3-13} \cline{15-25} \cline{27-37}

\multicolumn{37}{c}{} \\[-3ex] \cline{1-1} \cline{3-13} \cline{15-25} \cline{27-37}

3 &  &  &  & x & x &  &  &  & x &  & \textbf{0.2} & 0.49 &  & x &  & x &  &  &  &  & x &  & \textbf{0.25} & \textbf{0.57}  &  &  &  & x &  &  &  &  & x & x & 0.24 & \textbf{0.56}  \\ \cline{3-13} \cline{15-25} \cline{27-37}
 &  &  &  & x &  &  &  &  & x & x & 0.21 & 0.49 &  &  &  & x &  &  & x &  &  & x & 0.28 & 0.67  &  &  &  & x &  &  &  & x & x &  & 0.27 & 0.58  \\ \cline{3-13} \cline{15-25} \cline{27-37}
 &  & x &  & x &  &  &  & x &  &  & 0.29 & 0.5 &  & x &  & x &  &  &  &  &  & x & 0.28 & 0.68  &  &  &  & x & x &  &  & x &  &  & 0.27 & 0.6  \\ \cline{1-1} \cline{3-13} \cline{15-25} \cline{27-37}

\multicolumn{37}{c}{} \\[-3ex] \cline{1-1} \cline{3-13} \cline{15-25} \cline{27-37}

4 &  &  &  & x & x &  &  &  & x & x & 0.21 & \textbf{0.47} &  & x &  & x & x &  &  &  & x &  & 0.34 & 0.66  &  &  &  & x &  &  &  & x & x & x & 0.27 & 0.58  \\ \cline{3-13} \cline{15-25} \cline{27-37}
 &  &  &  & x & x &  &  & x & x &  & \textbf{0.2} & 0.5 &  & x &  & x &  &  & x &  & x &  & 0.3 & 0.69  &  &  &  & x & x &  &  & x & x &  & 0.27 & 0.59  \\ \cline{3-13} \cline{15-25} \cline{27-37}
 &  & x &  & x &  &  &  & x & x &  & 0.31 & 0.51 &  & x &  & x &  &  & x &  &  & x & 0.28 & 0.72  &  &  &  & x & x &  &  &  & x & x & 0.25 & 0.6  \\ \cline{1-1} \cline{3-13} \cline{15-25} \cline{27-37}

\multicolumn{37}{c}{} \\[-3ex] \cline{1-1} \cline{3-13} \cline{15-25} \cline{27-37}

5 &  & x &  & x & x &  &  & x & x &  & 0.25 & 0.48 &  & x &  & x & x &  & x &  & x &  & 0.36 & 0.79  &  &  &  & x & x &  &  & x & x & x & 0.26 & 0.59  \\ \cline{3-13} \cline{15-25} \cline{27-37}
 &  & x &  & x & x &  &  &  & x & x & 0.32 & 0.59 &  & x &  & x &  & x & x &  & x &  & 0.34 & 0.88  &  &  &  & x & x & x &  &  & x & x & \textbf{0.22} & 0.63  \\ \cline{3-13} \cline{15-25} \cline{27-37}
 &  &  & x & x & x &  &  &  & x & x & 0.25 & 0.64 &  & x &  & x & x &  & x &  &  & x & 0.34 & 0.88  &  & x &  & x & x &  &  &  & x & x & \textbf{0.22} & 0.64  \\ \cline{1-1} \cline{3-13} \cline{15-25} \cline{27-37}

\multicolumn{37}{c}{} \\[-3ex] \cline{1-1} \cline{3-13} \cline{15-25} \cline{27-37}

6 &  & x & x & x & x & x &  & x &  &  & 0.26 & 0.68 &  & x &  & x & x & x & x &  & x &  & 0.43 & 1.02  &  &  &  & x & x &  & x & x & x & x & 0.28 & 0.65  \\ \cline{3-13} \cline{15-25} \cline{27-37}
 &  &  & x & x & x & x &  & x & x &  & 0.27 & 0.69 &  & x &  & x & x &  & x &  & x & x & 0.43 & 1.04  &  & x &  & x &  &  & x & x & x & x & 0.28 & 0.65  \\ \cline{3-13} \cline{15-25} \cline{27-37}
 &  &  & x & x & x & x &  &  & x & x & 0.25 & 0.7 &  & x &  & x & x & x & x &  &  & x & 0.41 & 1.07  &  &  &  & x & x & x & x & x & x &  & 0.27 & 0.66  \\ \cline{1-1} \cline{3-13} \cline{15-25} \cline{27-37}

\multicolumn{37}{c}{} \\[-3ex] \cline{1-1} \cline{3-13} \cline{15-25} \cline{27-37}

7 &  & x &  & x & x &  & x & x & x & x & 0.26 & 0.8 &  & x &  & x & x & x & x &  & x & x & 0.51 & 1.11  &  & x &  & x & x &  & x & x & x & x & 0.28 & 0.65  \\ \cline{3-13} \cline{15-25} \cline{27-37}
 &  & x & x & x & x & x &  & x & x &  & 0.29 & 0.81 &  & x &  & x & x & x & x & x &  & x & 0.55 & 1.26  &  & x &  & x & x & x &  & x & x & x & 0.24 & 0.67  \\ \cline{3-13} \cline{15-25} \cline{27-37}
 &  &  &  & x & x & x & x & x & x & x & 0.22 & 0.83 &  & x & x & x & x & x & x &  &  & x & 0.68 & 1.32  &  & x &  & x & x & x & x & x & x &  & 0.27 & 0.67  \\ \cline{1-1} \cline{3-13} \cline{15-25} \cline{27-37}

\multicolumn{37}{c}{} \\[-3ex] \cline{1-1} \cline{3-13} \cline{15-25} \cline{27-37}

8 &  & x &  & x & x & x & x & x & x & x & 0.27 & 0.92 &  & x & x & x & x & x & x &  & x & x & 0.74 & 1.49  &  & x &  & x & x & x & x & x & x & x & 0.26 & 0.68  \\ \cline{3-13} \cline{15-25} \cline{27-37}
 &  & x & x & x & x & x & x &  & x & x & 0.31 & 0.93 &  & x &  & x & x & x & x & x & x & x & 0.64 & 1.51  &  & x & x & x & x & x &  & x & x & x & 0.26 & 0.81  \\ \cline{3-13} \cline{15-25} \cline{27-37}
 &  & x & x & x & x & x & x & x &  & x & 0.29 & 0.94 &  & x & x & x & x & x & x & x & x &  & 0.81 & 1.62  &  & x & x & x & x & x & x & x & x &  & 0.31 & 0.84  \\ \cline{1-1} \cline{3-13} \cline{15-25} \cline{27-37}

\multicolumn{37}{c}{} \\[-3ex] \cline{1-1} \cline{3-13} \cline{15-25} \cline{27-37}

9 &  & x & x & x & x & x & x & x & x & x & 0.3 & 0.94 &  & x & x & x & x & x & x & x & x & x & 0.84 & 1.96  &  & x & x & x & x & x & x & x & x & x & 0.31 & 0.85  \\ \cline{1-1} \cline{3-13} \cline{15-25} \cline{27-37}

\multicolumn{13}{c}{} \\

\end{tabular}

\end{center}
\caption{Cross-Eyed database, test set: Verification results for an
increasing number of fused comparators. The best combinations are chosen
based on the lowest FRR @ FAR=0.01\% of cross-spectral experiments.
The best result of each column is marked in bold. 
} \label{tab:crosseyed-results-fusion}
\end{table}
\normalsize

\setlength{\tabcolsep}{6pt}

\begin{figure}[htb]
\centering
        \includegraphics[width=0.95\textwidth]{FAFR_all_together_xeyed_shortname.png}
\caption{Cross-Eyed database, test set: cross-spectral FA/FR curves of the individual systems (left: with raw scores, middle: after z-score normalization, right: after mapping to log-likelihood ratios). Solid curves represent FR curves, and dashed curves represent FA curves. The 'fusion' curves on the center and right plots represent the fusion of SIFT+LBP+Resnet101 (see the main text for details). Best seen in colour and zoomed.
}
\label{fig:crosseyed-results-scores-FAFR}
\end{figure}

\subsection{Results: Fusion of Periocular Comparators}
\label{sect:crosseyed-fusion}

We then carry out fusion experiments using all the available
comparators, according to the fusion schemes presented in Section~\ref{sect:fusion-methods}.
We have tested all the possible fusion combinations.
Whenever training is needed
(i.e. to compute calibration weights, z-normalization, SVM, or Random Forest models),
the training set of the Cross-Eyed database is used.
%
%
In Figure~\ref{fig:crosseyed-results-fusion},
we show the best results obtained for an increasing number $M$ of combined comparators.
Following the protocol of Cross-Eyed 2016, the best
combinations are chosen based on the lowest cross-spectral FRR @ FAR=0.01\%.
Then, the corresponding EER of the chosen combinations is reported as well in Figure~\ref{fig:crosseyed-results-fusion}.
We use the two mentioned calibration possibilities of the fusion method (Figure~\ref{fig:fusion_model}):
$i$) the scores from all comparators are calibrated together ($N=M$ in Equation~\ref{eq:LLRfusion}),
or $ii$) the score of each comparator is calibrated separately ($N=1$) and
the resulting calibrated scores are summed.
These cases are shown in Figure~\ref{fig:crosseyed-results-fusion}
as `LLR' and `LLR (sum)', respectively.

As it can be observed, a substantial performance improvement can be obtained
when combining several comparators. The best cross-spectral performance
is obtained with a combination of
2 to 3 comparators. The FRR remains approximately constant until 5 comparators are combined, and then
it deteriorates when including more. The EER, nevertheless, deteriorates earlier.
We also observe that the probabilistic fusion method based on calibration
(LLR) outperform all the others, demonstrating its superiority.
This is more evident at low FAR, with a relative FRR
reduction of $\sim$47\% in comparison to using one comparator only.
It is also better if all scores are calibrated together, rather than calibrating them
individually and then summing them up (`LLR' vs `LLR (sum)').
%
%
Regarding the other fusion methods, the SVM with a linear or polynomial kernel stands out in comparison to the others.
The polynomial kernel shows equal or better performance in some cases, but such kernel
is much slower to train.
It is also worth noting that the simple average rule (AVG) provides similar performance than
trained approaches like the SVM, although it deteriorates quickly with the combination of more than 3 comparators.
On the other hand, the Random Forest approach performs among the worst, regardless of the number
of decision trees employed.

In Table~\ref{tab:crosseyed-results-fusion}, we show the comparators involved in the best fusion cases.
For the sake of space, we only provide results with a selection of fusion approaches, according to the observations made above when discussing Figure~\ref{fig:crosseyed-results-fusion}: the LLR method (best case), SVM linear (a good runner-up which is also faster to train than its polynomial counterpart), and AVG or AVERAGE (a simple approach that does not need training).
To allow a more comprehensive analysis, we also provide not only the best cases but also the second and third best combinations for a given number of comparators.
It can be seen that the best combinations for any given number of comparators always involve the
SIFT method.
The excellent accuracy of the SIFT comparator is not jeopardized by
the fusion with other comparators that have a performance one or two orders of
magnitude worse, but it is complemented to obtain even better cross-spectral
error rates, especially with trained approaches.
A careful look at the combinations of Table~\ref{tab:crosseyed-results-fusion} shows
that the CNN comparators are also chosen first for the fusion.
Together with SIFT,
they are the comparators with the best individual performance,
and they appear to be very complementary too.
However, it should not be taken as a general statement that the best fusion combination
always involves the best individual comparators.
Different fusion algorithms may lead to different results \cite{[FierrezJain06],[Fierrez05d]}.
For example, the best FRR with the simple average rule involves the SAFE comparator.
It is also worth noting that other comparators with worse individual performance
and not based on deep networks
(such as SAFE, LBP, or NTNU) are also selected
in combinations that have a performance nearly as good as the best cases.
At the same time, this shows the power of the fusion approaches employed,
and especially of the calibration method,
which are capable of reducing error rates substantially by fusion of comparators
with very heterogeneous performance and different feature representations.

To further illustrate the benefit of using calibrated scores, we plot in Figure~\ref{fig:crosseyed-results-scores-FAFR} the False Acceptance/False Rejection (FA/FR) curves of the individual systems. This is done using raw scores of each system (left), normalized scores using z-score normalization (center), and calibrated scores (right).
One selected fusion case of Table~\ref{tab:crosseyed-results-fusion} (best combination of three systems: SIFT+LBP+ResNet101) is also plotted using average of normalized scores (center) and score calibration (right).
It can be seen that the raw scores of each system lies in a different range, even if all comparators are expected to produce a score between 0 and 1 ($[-1,1]$ with SAFE).
After z-score normalization, the impostor score distributions become aligned to a certain degree, since such normalization converts them to zero mean and unit variance. Also, the extent to which the genuine distributions spread are indicative of the performance of each system (in order: SIFT (gray), DenseNet201 (black), ResNet101 (green), etc.). However, this cannot always be expected, since the fusion (blue thick curve) is situated between the curves of the individual systems involved due to scores being averaged. The EER of each system occurs as a different score value too.
Similar effects can be expected with other popular normalization techniques like max-min, tanh, etc. \citep{[Jain05]}.
%
%
%
When scores are normalized by calibration, two phenomena occur: $i$) the FA and FR curves cross at $\sim$0 score (the EER is always situated at this point), since a positive log-likelihood-ratio output supports the genuine (mated) decision, and a negative value the opposite; and $ii$) the spread and order of the curves are indicative of the performance of each system. For example, the SIFT curves (gray) have a smaller slope and reach higher log-likelihood-ratios (both positive and negative), due to this system being significantly better than the others (Table~\ref{tab:crosseyed-results}). The FA and FR curves of the other systems are then ordered (both in positive and negative sides):  DenseNet201 (black), ResNet101 (green), VGG-Face (blue), etc.
Furthermore, after the fusion (blue thick curves), the slope of the curves is even less, reaching even higher score values on both extremes. Given that the performance of the fusion is better than any of the other systems, both the genuine and impostor scores are pushed towards the extremes of the horizontal axis. This reflects the probabilistic meaning of calibrated scores, in the sense that a better performance translates to a reduced uncertainty via higher absolute score values.

%

\begin{table}[htb]
\scriptsize
\begin{center}
\begin{tabular}{p{0.07cm}p{0.07cm}p{0.07cm}p{0.07cm}p{0.07cm}p{0.07cm}|cc|cc|ccc|}

\multicolumn{13}{c}{} \\

\multicolumn{13}{c}{\textbf{CROSS-EYED DATABASE: Cross-spectral performance (VIS-NIR)}} \\ \hline


\multicolumn{6}{c}{} & \multicolumn{2}{c}{\textbf{Training set}} &
\multicolumn{2}{c}
{\textbf{Test set}} & \multicolumn{3}{c}{\textbf{Competition} \cite{[sequeira16crosseyed]}}\\
\cline{7-13}

\multicolumn{1}{p{0.12cm}}{\begin{turn}{90}approach
\end{turn}} & \multicolumn{1}{p{0.07cm}}{\begin{turn}{90}\textbf{safe}
\end{turn}} &
\multicolumn{1}{p{0.07cm}}{\begin{turn}{90}\textbf{gabor}
\end{turn}} & \multicolumn{1}{p{0.07cm}}{\begin{turn}{90}\textbf{sift}
\end{turn}} & \multicolumn{1}{p{0.07cm}}{\begin{turn}{90}\textbf{lbp}
\end{turn}} &
\multicolumn{1}{p{0.07cm}|}{\begin{turn}{90}\textbf{hog}
\end{turn}} & \textbf{EER} & \textbf{FRR}  & \textbf{EER} & \textbf{FRR}  & \textbf{EER} & \textbf{GF2} & \textbf{Rank} \\
\hline\hline

HH3 &  &  x &  &  x &  x & 4.5 & 16.77 & 4.86 & 24.59  & 6.02 & 11.42 & 3$^{rd}$ \\
\hline

HH2  & x & x &  & x & x & 3.02 & 12.63 & 4.51 & 19.75  & 5.24 & 9.14 & 2$^{nd}$ \\
\hline

HH1 & x & x & x & x & x & 0 & 0 & 0.28
&  0.83   & 0.29 & 0 & 1$^{st}$ \\
\hline

\end{tabular}
\end{center}
\caption{Comparison with results of the Cross-Eyed 2016 Competition \cite{[sequeira16crosseyed]}.
GF2 is the Generalized FRR (GFRR) at a Generalized FAR (GFAR) of 0.01\%.
The GFRR and GFAR are generalizations of the FRR and FAR to include
Failure to Acquire (FTA) and Failure to Enroll (FTE) rates, according to
ISO/IEC standards \cite{[ISO-IEC19795-1]}.
The ranking in the evaluation of the submitted approaches is also given.
For more information, refer to \cite{[sequeira16crosseyed]}.
}
\label{tab:crosseyed-results-competition}
\end{table}
\normalsize

\subsection{Results: Comparison with the Cross-Eyed 2016 competition}

Table~\ref{tab:crosseyed-results-competition} shows the
results of the submission of Halmstad University to the Cross-Eyed
2016 competition.
We provide both the results reported by the organizers \cite{[sequeira16crosseyed]},
and our own computations on the training
and test sets of the database using the executables submitted
and the protocol described in Section~\ref{sect:cross-eyed-dbprotocol}.
%
%
For the evaluation, only the SAFE, GABOR, SIFT, LBP, and HOG comparators
were available.
We contributed with three different fusion
combinations, named HH1, HH2, and HH3,
with the HH1 combination obtaining the first position
in the competition. 
%
Two key differences in the results reported in
Table~\ref{tab:crosseyed-results-competition}
in comparison with the present paper
are that in our executables:
$i$) the score of each comparator was calibrated separately,
and the resulting calibrated scores were summed up;
and $ii$)
%
the LBP and HOG comparators 
employed
the Euclidean distance
(which is the popular choice in the literature with these
methods, instead of $\chi^2$).
%
At the time of submission,
the test set had not been released,
so our decisions could only be based on the results
on the training set.
We observed that
the SIFT comparator already provided
cross-spectral error rates of nearly 0\% on the training set
(not shown in Table~\ref{tab:crosseyed-results-competition}).
However, it was reasonable to expect a higher
error with a bigger dataset, as demonstrated later when the
test set was released. 
Therefore, we contributed to the
competition with a fusion of the five comparators available (called HH1)
to be able to better cope with the generalization issue that is expected
when performance is measured in a bigger set of images.
Indeed, in Table~\ref{tab:crosseyed-results-competition} it can be seen
that performance on the test set is systematically worse than on the
training set.
Since the combination of the five available comparators is
computationally heavy in template size (due to the SIFT comparator), we
also contributed by removing SIFT (combination HH2), and by further
removing SAFE (combination HH3),
which has a feature extraction time
considerably higher than the rest of the comparators in our implementation
(see Table~\ref{tab:file-size-time}).
Thus, our motivation behind HH2 and HH3 was to reduce template size
and feature extraction time.
Some differences are observable between our results with the test
set and the results reported by the competition
\cite{[sequeira16crosseyed]}. We attribute this to two factors:
$i$) the additional 10 subjects included in the test set released,
which were not used during the competition, 
and $ii$) the
employment of a different test protocol since it is not
specified by the organizers the exact images used for impostor
trials during the competition.
Therefore, the experimental framework used in this paper is not
exactly the same employed in the Cross-Eyed competition.

\section{Cross-Sensor (VIS-VIS) Smartphone Periocular Recognition}
\label{sect:vssiris}

\subsection{Database and Protocol}

In the cross-sensor experiments of this section,
we use the Visible Spectrum
Smartphone Iris (VSSIRIS) database \cite{[Raja14b]}, which has
images from 28 subjects (56 eyes) captured using the rear camera of
two smartphones (Apple iPhone 5S, of 3264$\times$2448 pixels, and
Nokia Lumia 1020, of 3072$\times$1728 pixels).
They have been obtained in unconstrained conditions under mixed
illumination (natural sunlight and artificial room light).
Each eye has 5 samples per smartphone, thus
5$\times$56=280 images per device (560 in total). The acquisition is
made without flash, in a single session and with semi-cooperative
subjects.
Figure~\ref{fig:db-samples} (bottom) shows some examples.

All images of VSSIRIS are annotated manually, so the radius and centre
of the pupil and sclera circles are available. 
%
Images are resized via bicubic interpolation to have the same sclera
radius (set to $R_s$=145, the average radius of the whole database).
We use the sclera for normalization since it is not affected by
dilation.
Then, images are aligned by extracting a square region of
$6R_s$$\times$$6R_s$ (871$\times$871) around the sclera centre. This
size is set empirically to ensure that all available images have
sufficient margin to the four sides of the sclera centre.
Here, there is sufficient availability to the four sides of the eye,
so the normalized images have the eye centred in the image, as can
be seen in Figure~\ref{fig:img-ROI} (bottom).
Images are further processed by Contrast-Limited Adaptive Histogram
Equalization (CLAHE) \cite{[Zuiderveld94clahe]} to compensate for
variability in local illumination. 

\begin{table}[htb]
\small
\begin{center}
\begin{tabular}{ccc}

 \multicolumn{3}{c}{} \\

 \multicolumn{3}{c}{\textbf{VSSIRIS database}} \\ \hline

 \multicolumn{1}{c}{\textbf{Protocol}}&
 \multicolumn{1}{c}{} &  \multicolumn{1}{c}{} \\

 \multicolumn{1}{c}{(28 subjects)} &
 \multicolumn{1}{c}{Same-Sensor} &
 \multicolumn{1}{c}{Cross-Sensor}
\\ \hline \hline

Genuine  & 56 $\times$ (4+3+2+1) = 560 & 56 $\times$ 5 $\times$ 5 = 1,400  \\

Impostor  & 56 $\times$ 55 = 3,080  & 56$ \times$ 55 = 3,080   \\
\hline \hline

\end{tabular}

\end{center}
\caption{VSSIRIS database: Experimental protocol.}
\label{tab:exp-protocol-vssiris}
\end{table}
\normalsize

We carry out verification experiments, with each eye considered a
different user. We compare images both from the same device
(\emph{same-sensor}) and from different devices
(\emph{cross-sensor}). Genuine trials are obtained by comparing
each image of an eye to the remaining images of the same eye,
avoiding symmetric comparisons. Impostor trials are done by comparing
the 1$^{st}$ image of an eye to the 2$^{nd}$ image of the remaining
eyes. The experimental protocol is summarized in
Table~\ref{tab:exp-protocol-vssiris}.
The smaller size of VSSIRIS in comparison with the Cross-Eyed
database results in the availability of fewer scores.
Therefore, whenever a parameter needs training,
2-fold cross-validation \cite{[Jain00b]} is used,
dividing the available number of users in two partitions.
Otherwise, we report results employing the entire VSSIRIS database.
%
%
%

The parameters of the periocular comparators are as follows.
As with the Cross-Eyed database, they are designed to adapt dynamically to
the size of the image, being the sclera boundary the only necessary input.
Regarding the SAFE comparator, the annular band of the first
circular ring starts at the sclera circle ($R$=145 pixels), and the
band of the last ring ends at the boundary of the image, resulting
in a ROI of 871$\times$871 pixels around the eye centre.
The availability of sufficient margin around the four sides of the
eye makes possible to have a bigger ROI with VSSIRIS, as can be
shown in Figure~\ref{fig:img-ROI}, third column.
This availability also allows one extra row in the grid employed
with GABOR, LBP and HOG comparators, having 8$\times$8=64 non-overlapping blocks. Given the size of the input image, each block has 109$\times$109 pixels.
%
%
For consistency with Cross-Eyed, the eight blocks of the image centre are not considered, effectively resulting in 56 blocks (some more than Cross-Eyed, which has 48 blocks
of size 88$\times$88 each).
The GABOR comparator employs filter wavelengths spanning from 55 to 7
pixels, which are set proportional to the block size as
109/2$\approx$55 to 109/16$\approx$7.
Regarding VGG-Face, Resnet101
and Densenet201, images are resized to 224$\times$224,
which are the input dimensions of these CNNs.
With SIFT, we keep as baseline the use of the annular ROI, but for comparison purposes, we also evaluate the entire input image (both at the original size of 871$\times$871 and at 224$\times$224).
Table~\ref{tab:vector-size} (third column) indicates the size of the feature vector
for a given periocular image with the different comparators employed.
The full ROI produces an average of 3824 SIFT key-points (489472 values), and $\sim$3000 with the annular ROI (384000 values), which are higher values than Cross-Eyed, since the image is bigger. At 224$\times$224, there are 130 key-points per image on average (16512 values).
Experiments have been done in the same machine and with the same
algorithm implementations than Cross-Eyed
(Section~\ref{sect:cross-eyed-dbprotocol}).
%
%
The feature extraction and comparison
times are given in Table~\ref{tab:file-size-time} (right).

\begin{figure}[htb]
\centering
        \includegraphics[width=0.95\textwidth]{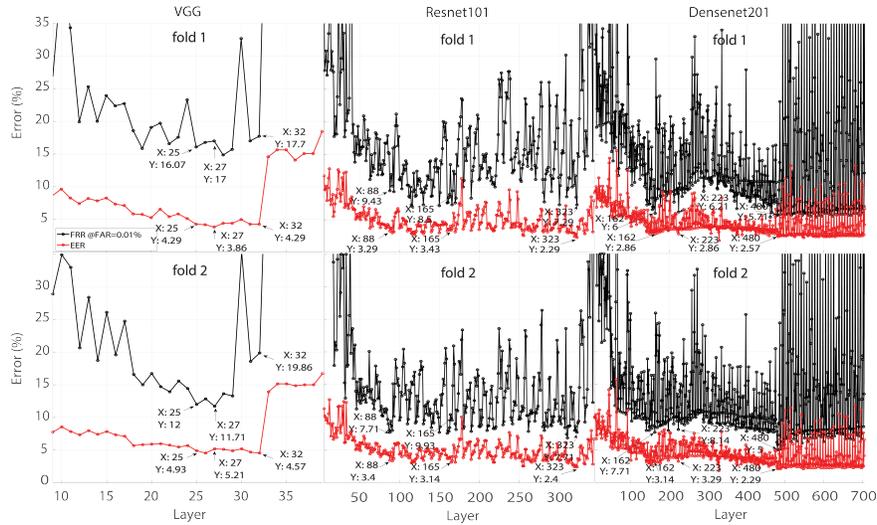}
\caption{VSSIRIS database: Cross-sensor accuracy (VIS-VIS) of different CNN layers.}
\label{fig:vssiris-results-cnns}
\end{figure}

\subsection{Results: Finding the Optimum Layer of the Convolutional Neural Networks}

We first identify the optimum layer of each CNN. The cross-sensor accuracy of each network
is given in Figure~\ref{fig:vssiris-results-cnns} for each cross-validation fold.
When selecting the best layer, we have tried to find the one that
gives optimum performance both
with the Cross-Eyed and the VSSIRIS databases simultaneously.
However, it has not always been possible.
According to the discussion in Section~\ref{subsec:best-CNN-layer-crosseyed}, the best
layers with VSSIRIS are layer 25 (VGG-Face), layer 323 (ResNet101), and layer 480 (DenseNet201).
It can be seen as well that the optimum layers of VSSIRIS are the same for the two folds.
With Cross-Eyed, on the other hand, the best layers were not so deep: 165 (ResNet101) and layer 223 (DenseNet201).


\begin{figure}[htb]
\centering
\includegraphics[width=0.45\textwidth]{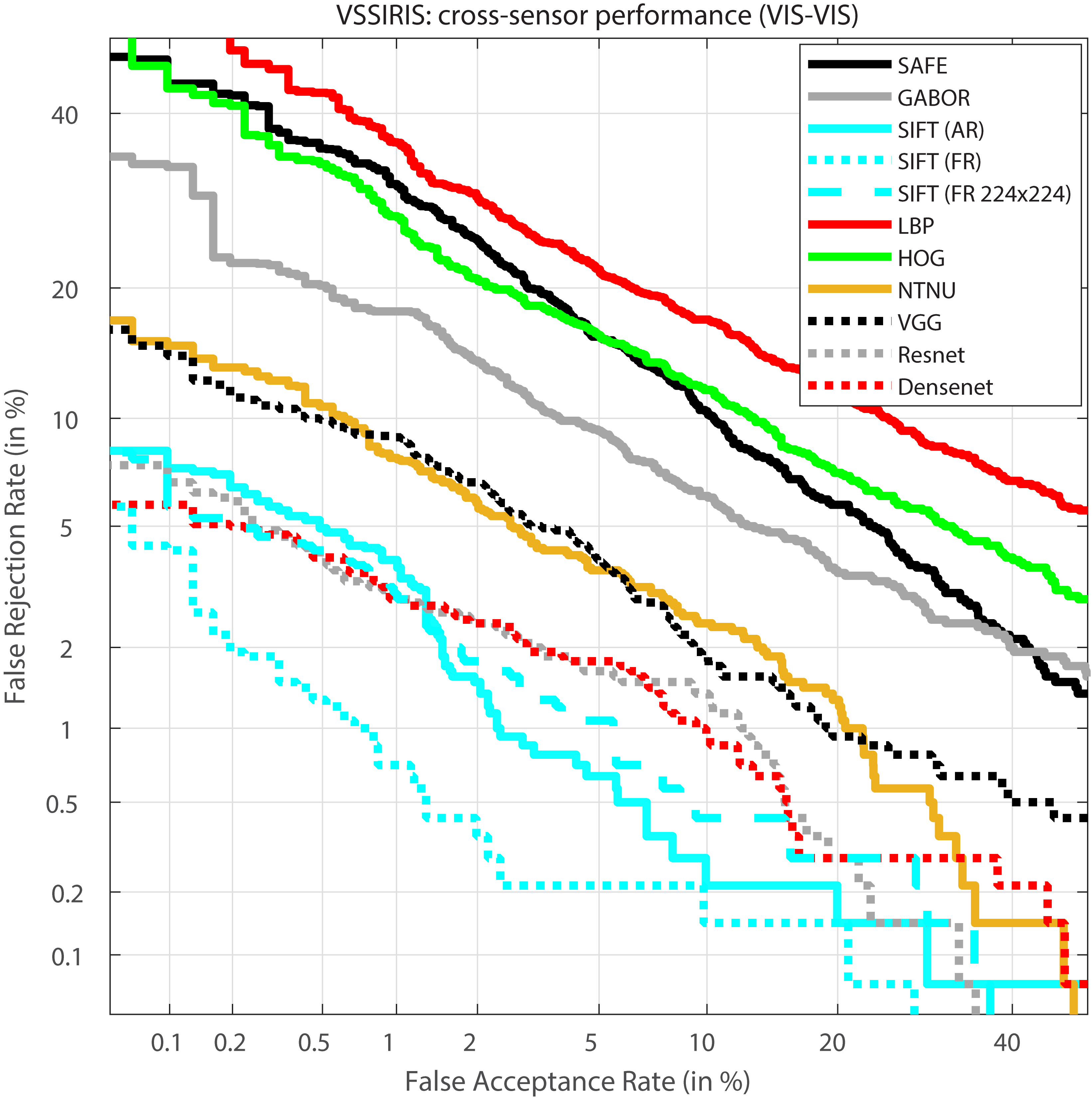}
\caption{VSSIRIS database: Verification results of the individual
comparators. Best seen in colour.} \label{fig:vssiris-results1}
\end{figure}

\begin{table}[htb]
\scriptsize
\begin{center}
\begin{tabular}{ccccccccc}

\multicolumn{9}{c}{\textbf{}} \\

 \multicolumn{2}{c}{} &  \multicolumn{3}{c}{\textbf{Equal Error Rate (EER)}}
 & \multicolumn{1}{c}{} &  \multicolumn{3}{c}{\textbf{FRR @
 FAR=0.01\%}}\\ \cline{3-5} \cline{7-9}

 \multicolumn{1}{c}{\textbf{}} & \multicolumn{1}{c}{\textbf{}} &
\multicolumn{2}{c}{\textbf{Same sensor}}
& \multicolumn{1}{c}{\textbf{}}
& \multicolumn{1}{c}{\textbf{}} & \multicolumn{2}{c}{\textbf{Same sensor}}  & \multicolumn{1}{c}{\textbf{}}\\
 \cline{3-4} \cline{7-8}

\multicolumn{1}{c}{\textbf{comparator}} & \multicolumn{1}{c}{\textbf{}} &
 \multicolumn{1}{c}{\textbf{iPhone}} &
 \multicolumn{1}{c}{\textbf{Nokia}} &
 \multicolumn{1}{c}{\textbf{cross-sensor}} & &
 \multicolumn{1}{c}{\textbf{iPhone}} &
 \multicolumn{1}{c}{\textbf{Nokia}} &
 \multicolumn{1}{c}{\textbf{cross-sensor}}
 \\ \cline{1-1} \cline{3-5} \cline{7-9}

SAFE & & 1.6\%  & 2.6\% & 10.2\% (+537.5\%) & & 4.6\% & 11.1\% & 50.9\% (+1006.5\%)  \\
\cline{1-1} \cline{3-5} \cline{7-9}

GABOR & & 2.1\%  & 1.5\% & 7.3\% (+386.7\%) &  & 4.3\% & 8.9\% & 39.1\% (+809.3\%)  \\
\cline{1-1} \cline{3-5} \cline{7-9}

SIFT (AR 871$\times$871) & & 0\% & 0.1\% & 1.6\% ($>$1500\%)  &  & 0\% & 0.7\% & 12.7\% ($>$1700\%)  \\
\cline{1-1} \cline{3-5} \cline{7-9}

SIFT (FR 871$\times$871) & & 0\% & 0\% & 0.82\% (-)  &  & 0\% & 0\% & 6.25\% (-)  \\
\cline{1-1} \cline{3-5} \cline{7-9}

SIFT (FR 224$\times$224) & & 0\% & 0\% & 1.79\% (-)  &  & 0\% & 0\% & 10.54\% (-)  \\
\cline{1-1} \cline{3-5} \cline{7-9}



LBP & & 4.8\% & 4.9\% & 14.1\% (+193.8\%) &  & 6.8\% & 16.8\% & 71.2\% (+947.1\%)  \\
\cline{1-1} \cline{3-5} \cline{7-9}

HOG & & 3.9\% & 4.5\% & 11\% (+182.1\%) &  & 5.2\% & 17.3\% & 70.7\% (+1259.6\%)  \\
\cline{1-1} \cline{3-5} \cline{7-9}

NTNU & & 0.7\% & 0.7\% & 4.1\% (+480\%) &  & 0.9\% & 1.8\% & 23.1\% (+2500\%)  \\
\cline{1-1} \cline{3-5} \cline{7-9}

VGG-Face & & 0.9\% & 0.7\% & 4.4\% (+528.6\%) &  & 1.6\% & 1.3\% & 20.8\% (+1500\%)  \\
\cline{1-1} \cline{3-5} \cline{7-9}

Resnet101 & & 0.5\% & 0\% & 2.3\% (-) &  & 0.7\% & 0.4\% & 10.3\% (+2475\%)
 \\ \cline{1-1} \cline{3-5} \cline{7-9}

Densenet201 & & 0.5\% & 0\% & 2.4\% (-) &  & 0.7\% & 0.2\% & 6.2\% (+3000\%)
 \\ \cline{1-1} \cline{3-5} \cline{7-9}

\end{tabular}

\end{center}
\caption{VSSIRIS database: Verification results of the individual
comparators. The relative variation of cross-sensor performance with
respect to the best same-sensor performance is given in brackets (for the SIFT comparator, the result is calculated w.r.t. the Nokia performance, since the iPhone performance is 0\%, which would result in Inf due to division by zero; if both Nokia and iPhone performance is 0\%, no value is given). AR=annular ROI. FR=Full ROI.}
\label{tab:vssiris-results}
\end{table}
\normalsize

\subsection{Results: Individual Comparators}

The performance of individual comparators is then reported in Table~\ref{tab:vssiris-results}.
%
%
Similarly as Section~\ref{sect:cross-eyed},
we adopt as measures of accuracy
the EER and the FRR at FAR=0.01\%.
In Figure~\ref{fig:vssiris-results1}, we give the DET curves of the cross-sensor experiments.
%

%
By comparing Table~\ref{tab:crosseyed-results} and
Table~\ref{tab:vssiris-results}, it can be observed that same-sensor
experiments with the VSSIRIS database usually exhibit lower error rates
for any given comparator.
Possible explanations might be that the ROI of
VSSIRIS images is bigger (871$\times$871 vs 613$\times$701),
or that the VSSIRIS database has fewer users
(28 vs 90 subjects).
%
%
On the opposite side, cross-sensor error rates with VSSIRIS
are significantly worse for some comparators (e.g. SIFT, HOG, NTNU, or VGG-Face).
\textit{Lighter} comparators such as LBP or HOG are not capable of providing good
cross-sensor performance in low-security applications either (EER of 11\% or higher).
%
%
%
%
The difference is
especially relevant with the SIFT comparator, where cross-sensor error
rates on Cross-Eyed (Table~\ref{tab:crosseyed-results})
were 0.28\% (EER) and 0.88\% (FRR),
but here they increase one order of magnitude, up to 1.6\% (EER) and 12.7\% (FRR) (annular ROI).
This is despite the higher number of SIFT key-points per image with VSSIRIS
due to higher image size ($\sim$3000 vs $\sim$1900 on average).
%
%
%
It is thus interesting that the comparators employed in this paper are more robust to
the variability between images in different spectra (NIR and VIS)
than the variability between images in the same (VIS) spectrum
captured with two different smartphones. 
Such effect can also be seen in that the SIFT comparator is more sensitive to changes in the ROI with VSSIRIS. With the full ROI (FR), cross-sensor errors are divided by two, so a bigger ROI can be seen as a way to counteract cross-sensor variability in this case.
Another difference here is that if the image size is reduced to 224$\times$224, SIFT does not degrade as much, being in some regions of the DET at the same level than the baseline annular ROI (AR) or than some CNNs.
It should be noted, though, that images in Cross-Eyed are obtained
with a dual spectrum sensor, which captures NIR and VIS images
synchronously. 
Thus,
in practice, there is no scale, 3D rotation or time-lapse difference between
corresponding NIR and VIS samples.
Only a spatial offset between the two exist in the plane perpendicular to the optical axes of the cameras due to the sensors not being perfectly calibrated (which can be noticed in Figure~\ref{fig:db-samples}), so images are expected to be very well aligned after cropping.
This synchronicity and absence of time span
could be one of the reasons of
the better cross-spectral performance obtained with the Cross-Eyed database, or the less sensitivity of the SIFT method to changes in the ROI.
%
%

Another observation is that
%
same-sensor performance with VSSIRIS is sometimes very
different depending on the smartphone employed,
even if they involve the same
subjects and images are resized to the same size. Contrarily,
same-sensor performance with Cross-Eyed tends to be similar
regardless of the spectrum employed
(Table~\ref{tab:crosseyed-results}), which might be explained
as well by the synchronicity in the acquisition mentioned above.
Previous works have suggested that discrepancy in colours between VIS
sensors can lead to variability in performance, which is further
amplified when images from such sensors are compared among them.
The sensitivity of SIFT to changes in the ROI can also be an indicative of this.
Although we apply local adaptive contrast equalization, our results
suggest that other device-dependent colour correction might be of
help 
\cite{[Santos14]}.
%
%
Another difference observed here is that the best individual comparator (in terms of FRR) is not SIFT. With Cross-Eyed, SIFT was the best by a large margin, but here, other comparators have similar or better performance (e.g. DenseNet201, ResNet101).
This is despite the higher number of SIFT key-points per image with VSSIRIS mentioned above.
Nevertheless, the correlation between bigger template size and lower error rates remains
since the comparators with the best performance (SIFT, NTNU and the three CNNs) are
also the ones with the biggest feature vector.
The superiority of these comparators can also be observed in the DET curves of Figure~\ref{fig:vssiris-results1}.

\begin{figure}[htb]
\centering
        \includegraphics[width=0.95\textwidth]{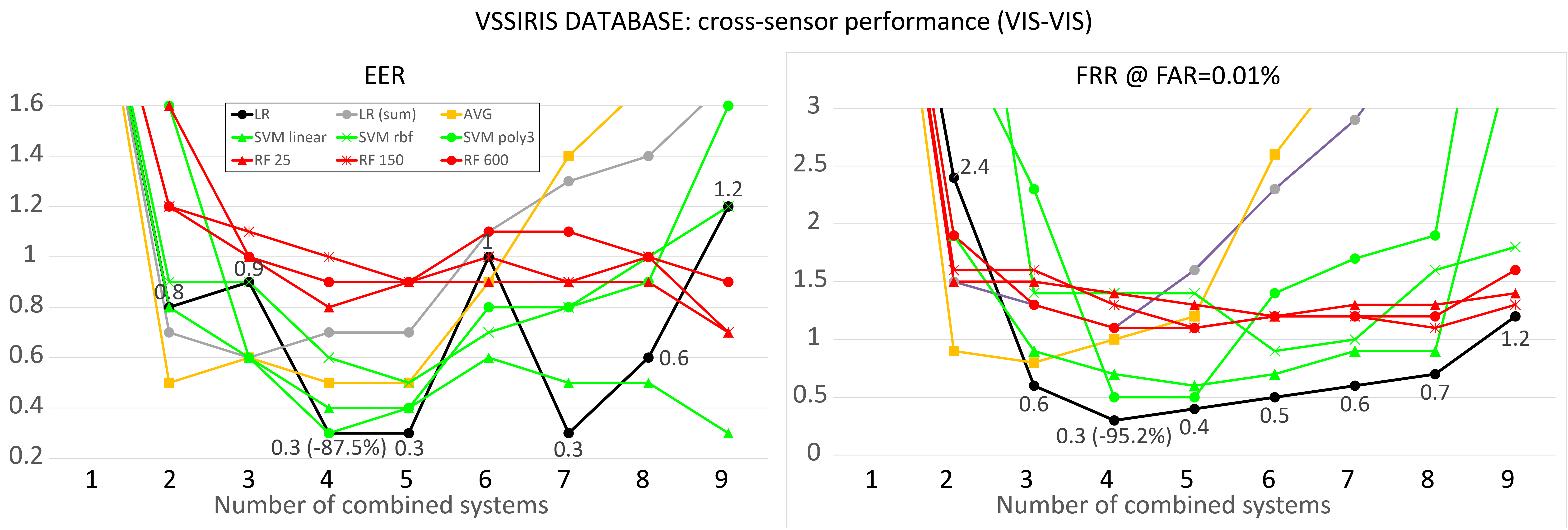}
\caption{VSSIRIS database, test set: Verification results for an increasing number of fused comparators. Best seen in colour.}
\label{fig:vssiris-results-fusion}
\end{figure}

\subsection{Results: Fusion of Periocular Comparators}
\label{sect:vissiris-fusion}

We now carry out fusion experiments using all the available comparators.
Whenever a fusion method needs training,
2-fold cross-validation \cite{[Jain00b]} was used,
dividing the available number of users in two partitions.
We have also tested here all the possible fusion combinations,
with the best combinations chosen based on the lowest cross-sensor
FRR @ FAR=0.01\%.
The best results obtained for an increasing number $M$ of combined comparators
is given in Figure~\ref{fig:vssiris-results-fusion}
(average values of the two folds).
The comparators involved in the best fusion cases are also given
in Table~\ref{tab:vssiris-results-fusion}
(as in Section~\ref{sect:crosseyed-fusion},
the table only shows the results of a selection of fusion approaches).


\setlength{\tabcolsep}{0pt}

\begin{table}[htb]
\tiny
\begin{center}

\begin{tabular}{|P{0.25cm}|P{0.2cm}|P{0.2cm}P{0.2cm}P{0.2cm}P{0.2cm}P{0.2cm}P{0.2cm}P{0.2cm}P{0.2cm}P{0.2cm}|P{0.7cm}|P{0.7cm}|p{0.2cm}|P{0.2cm}P{0.2cm}P{0.2cm}P{0.2cm}P{0.2cm}P{0.2cm}P{0.2cm}P{0.2cm}P{0.2cm}|P{0.7cm}|P{0.7cm}|p{0.2cm}|P{0.2cm}P{0.2cm}P{0.2cm}P{0.2cm}P{0.2cm}P{0.2cm}P{0.2cm}P{0.2cm}P{0.2cm}|P{0.7cm}|P{0.7cm}|}

\multicolumn{37}{c}{\textbf{VSSIRIS DATABASE: cross-sensor performance (VIS-VIS)}} \\
\cline{3-13} \cline{15-25} \cline{27-37}

\multicolumn{1}{c}{} & \multicolumn{1}{c}{} & \multicolumn{11}{|c|}{\textbf{LLR FUSION}} & \multicolumn{1}{c}{} & \multicolumn{11}{|c|}{\textbf{AVERAGE FUSION}} & \multicolumn{1}{c}{} & \multicolumn{11}{|c|}{\textbf{SVM LINEAR FUSION}}  \\  \cline{3-13} \cline{15-25} \cline{27-37}


\multicolumn{1}{P{0.12cm}}{\begin{turn}{90}\textbf{\# comparator}
\end{turn}} & \multicolumn{1}{c}{} &
\multicolumn{1}{|P{0.04cm}}{\begin{turn}{90}\textbf{safe}
\end{turn}} &
\multicolumn{1}{P{0.04cm}}{\begin{turn}{90}\textbf{gabor}
\end{turn}} & \multicolumn{1}{P{0.04cm}}{\begin{turn}{90}\textbf{sift}
\end{turn}} & \multicolumn{1}{P{0.04cm}}{\begin{turn}{90}\textbf{lbp}
\end{turn}} &
\multicolumn{1}{P{0.04cm}}{\begin{turn}{90}\textbf{hog}
\end{turn}} & \multicolumn{1}{P{0.04cm}}{\begin{turn}{90}\textbf{ntnu}
\end{turn}} & \multicolumn{1}{P{0.04cm}}{\begin{turn}{90}\textbf{vgg-face}
\end{turn}} & \multicolumn{1}{P{0.04cm}}{\begin{turn}{90}\textbf{resnet101}
\end{turn}} & \multicolumn{1}{P{0.04cm}|}{\begin{turn}{90}\textbf{densenet201}
\end{turn}}
& EER (\%) & FRR (\%) &
\multicolumn{1}{P{0.12cm}}{} & \multicolumn{1}{|P{0.04cm}}{\begin{turn}{90}\textbf{safe}
\end{turn}} &
\multicolumn{1}{P{0.04cm}}{\begin{turn}{90}\textbf{gabor}
\end{turn}} & \multicolumn{1}{P{0.04cm}}{\begin{turn}{90}\textbf{sift}
\end{turn}} & \multicolumn{1}{P{0.04cm}}{\begin{turn}{90}\textbf{lbp}
\end{turn}} &
\multicolumn{1}{P{0.04cm}}{\begin{turn}{90}\textbf{hog}
\end{turn}} & \multicolumn{1}{P{0.04cm}}{\begin{turn}{90}\textbf{ntnu}
\end{turn}} & \multicolumn{1}{P{0.04cm}}{\begin{turn}{90}\textbf{vgg-face}
\end{turn}} & \multicolumn{1}{P{0.04cm}}{\begin{turn}{90}\textbf{resnet101}
\end{turn}} & \multicolumn{1}{P{0.04cm}|}{\begin{turn}{90}\textbf{densenet201}
\end{turn}}
& EER (\%) & FRR (\%) &
\multicolumn{1}{P{0.12cm}}{} & \multicolumn{1}{|P{0.04cm}}{\begin{turn}{90}\textbf{safe}
\end{turn}} &
\multicolumn{1}{P{0.04cm}}{\begin{turn}{90}\textbf{gabor}
\end{turn}} & \multicolumn{1}{P{0.04cm}}{\begin{turn}{90}\textbf{sift}
\end{turn}} & \multicolumn{1}{P{0.04cm}}{\begin{turn}{90}\textbf{lbp}
\end{turn}} &
\multicolumn{1}{P{0.04cm}}{\begin{turn}{90}\textbf{hog}
\end{turn}} & \multicolumn{1}{P{0.04cm}}{\begin{turn}{90}\textbf{ntnu}
\end{turn}} & \multicolumn{1}{P{0.04cm}}{\begin{turn}{90}\textbf{vgg-face}
\end{turn}} & \multicolumn{1}{P{0.04cm}}{\begin{turn}{90}\textbf{resnet101}
\end{turn}} & \multicolumn{1}{P{0.04cm}|}{\begin{turn}{90}\textbf{densenet201}
\end{turn}}
& EER (\%) & FRR (\%) \\
\cline{1-1} \cline{3-13} \cline{15-25} \cline{27-37}

1  &  &  &  &  &  &  &  &  &  & x & 2.4 & 6.2   &  &  &  &  &  &  &  &  &  & x & 2.4 & 6.2   &  &  &  &  &  &  &  &  &  & x & 2.4 & 6.2  \\ \cline{3-13} \cline{15-25} \cline{27-37}
  &  &  &  &  &  &  &  &  & x &  & 2.3 & 10.3   &  &  &  &  &  &  &  &  & x &  & 2.3 & 10.3   &  &  &  &  &  &  &  &  & x &  & 2.3 & 10.3   \\ \cline{3-13} \cline{15-25} \cline{27-37}
 &  &  &  & x &  &  &  &  &  &  & 1.6 & 12.7   &  &  &  & x &  &  &  &  &  &  & 1.6 & 12.7   &  &  &  & x &  &  &  &  &  &  & 1.6 & 12.7    \\
 \cline{1-1} \cline{3-13} \cline{15-25} \cline{27-37}

\multicolumn{37}{c}{} \\[-3ex] \cline{1-1} \cline{3-13} \cline{15-25} \cline{27-37}

2  &  &  &  & x &  &  &  &  &  & x & 0.8 & 2.4   &  &  &  & x &  &  &  &  &  & x & \textbf{0.5} & 0.9   &  &  &  & x &  &  &  &  & x &  & 0.8 & 1.9  \\ \cline{3-13} \cline{15-25} \cline{27-37}
  &  &  &  & x &  &  &  &  & x &  & 0.9 & 2.4   &  &  &  & x &  &  &  &  & x &  & 0.6 & 1.6   &  &  &  & x &  &  &  &  &  & x & 0.7 & 1.9  \\ \cline{3-13} \cline{15-25} \cline{27-37}
  &  &  &  & x &  &  &  & x &  &  & 1 & 2.8   &  &  &  & x &  &  &  & x &  &  & 0.9 & 2.8   &  &  &  & x &  &  &  & x &  &  & 0.9 & 2.7  \\ \cline{1-1} \cline{3-13} \cline{15-25} \cline{27-37}

\multicolumn{37}{c}{} \\[-3ex] \cline{1-1} \cline{3-13} \cline{15-25} \cline{27-37}

3  &  & x &  & x &  &  &  &  &  & x & 0.9 & 0.6   &  &  &  & x &  &  & x &  &  & x & 0.6 & \textbf{0.8}   &  & x &  & x &  &  &  &  &  & x & 0.6 & 0.9    \\ \cline{3-13} \cline{15-25} \cline{27-37}
  &  &  &  & x &  &  & x &  &  & x & 1.4 & 0.6   &  & x &  & x &  &  &  &  &  & x & \textbf{0.5} & 0.9   &  &  & x & x &  &  &  &  &  & x & 0.6 & 1   \\ \cline{3-13} \cline{15-25} \cline{27-37}
  &  &  &  & x & x &  &  &  &  & x & \textbf{0.3} & 0.7   &  &  &  & x &  &  &  &  & x & x & \textbf{0.5} & 1.1   &  &  &  & x & x &  &  &  & x &  & 0.5 & 1    \\ \cline{1-1} \cline{3-13} \cline{15-25} \cline{27-37}

\multicolumn{37}{c}{} \\[-3ex] \cline{1-1} \cline{3-13} \cline{15-25} \cline{27-37}

4  &  & x &  & x & x &  &  &  &  & x & \textbf{0.3} & \textbf{0.3}   &  & x &  & x &  &  & x &  &  & x & \textbf{0.5} & 1   &  & x & x & x &  &  &  &  &  & x & 0.4 & 0.7    \\ \cline{3-13} \cline{15-25} \cline{27-37}
  &  & x &  & x &  &  &  & x & x &  & 0.6 & 0.5   &  & x &  & x &  &  & x &  & x &  & 0.7 & 1   &  & x &  & x &  & x &  &  &  & x & 0.6 & 0.8  \\ \cline{3-13} \cline{15-25} \cline{27-37}
  &  & x & x & x &  &  &  &  &  & x & 0.5 & 0.5   &  & x &  & x &  &  &  &  & x & x & \textbf{0.5} & 1.1   &  & x &  & x & x &  &  &  &  & x & \textbf{0.2} & 0.8   \\ \cline{1-1} \cline{3-13} \cline{15-25} \cline{27-37}

\multicolumn{37}{c}{} \\[-3ex] \cline{1-1} \cline{3-13} \cline{15-25} \cline{27-37}

5  &  & x &  & x & x &  &  &  & x & x & \textbf{0.3} & 0.4   &  & x &  & x &  &  & x &  & x & x & \textbf{0.5} & 1.2   &  & x & x & x &  & x &  &  &  & x & 0.4 & \textbf{0.6}  \\ \cline{3-13} \cline{15-25} \cline{27-37}
  &  & x &  & x &  &  &  & x & x & x & 0.5 & 0.4   &  & x &  & x &  &  &  & x & x & x & 0.9 & 2.3   &  & x &  & x & x &  &  &  & x & x & 0.3 & 0.7   \\ \cline{3-13} \cline{15-25} \cline{27-37}
  &  &  &  & x &  &  & x & x & x & x & 1.4 & 0.6   &  & x &  & x & x &  &  &  & x & x & 1.1 & 2.4   &  & x &  & x &  & x & x &  &  & x & 0.6 & 0.8   \\ \cline{1-1} \cline{3-13} \cline{15-25} \cline{27-37}

\multicolumn{37}{c}{} \\[-3ex] \cline{1-1} \cline{3-13} \cline{15-25} \cline{27-37}

6  &  & x &  & x &  &  & x & x & x & x & 1 & 0.5   &  & x &  & x &  &  & x & x & x & x & 0.9 & 2.6   &  & x &  & x &  & x & x &  & x & x & 0.6 & 0.7    \\ \cline{3-13} \cline{15-25} \cline{27-37}
  &  & x &  & x &  & x &  & x & x & x & \textbf{0.3} & 0.5   &  & x &  & x &  & x & x &  & x & x & 1 & 2.7   &  & x & x & x &  & x &  &  & x & x & 0.5 & 0.8  \\ \cline{3-13} \cline{15-25} \cline{27-37}
 &  & x &  & x & x &  &  & x & x & x & \textbf{0.3} & 0.5   &  & x &  & x & x &  & x &  & x & x & 1 & 3   &  & x & x & x &  & x &  & x &  & x & 0.6 & 0.8  \\ \cline{1-1} \cline{3-13} \cline{15-25} \cline{27-37}

\multicolumn{37}{c}{} \\[-3ex] \cline{1-1} \cline{3-13} \cline{15-25} \cline{27-37}

7  &  & x & x & x & x & x &  & x &  & x & \textbf{0.3} & 0.6   &  & x & x & x &  & x & x &  & x & x & 1.4 & 3.5   &  & x & x & x &  &  & x & x & x & x & 0.5 & 0.9   \\ \cline{3-13} \cline{15-25} \cline{27-37}
  &  & x & x & x & x & x & x &  & x &  & 1 & 0.6   &  & x &  & x &  & x & x & x & x & x & 1.4 & 3.6   &  & x & x & x &  & x &  & x & x & x & 0.6 & 0.9  \\ \cline{3-13} \cline{15-25} \cline{27-37}
 &  & x & x & x & x &  &  & x & x & x & \textbf{0.3} & 0.7   &  & x &  & x & x & x & x &  & x & x & 1.2 & 3.8   &  & x & x & x &  & x & x &  & x & x & 0.6 & 0.9  \\ \cline{1-1} \cline{3-13} \cline{15-25} \cline{27-37}

\multicolumn{37}{c}{} \\[-3ex] \cline{1-1} \cline{3-13} \cline{15-25} \cline{27-37}

8  &  & x & x & x &  & x & x & x & x & x & 0.6 & 0.7   &  & x & x & x &  & x & x & x & x & x & 1.7 & 4.1   &  & x & x & x &  & x & x & x & x & x & 0.5 & 0.9   \\ \cline{3-13} \cline{15-25} \cline{27-37}
  &  & x & x & x & x & x & x & x & x &  & 1.1 & 0.8   &  & x & x & x & x & x & x &  & x & x & 1.8 & 4.2   &  & x & x & x & x & x &  & x & x & x & \textbf{0.2} & 1.8    \\ \cline{3-13} \cline{15-25} \cline{27-37}
  &  & x & x & x & x & x & x &  & x & x & 0.9 & 0.9   &  & x &  & x & x & x & x & x & x & x & 1.6 & 4.5   &  & x & x & x & x & x & x & x &  & x & 0.3 & 2.4  \\ \cline{1-1} \cline{3-13} \cline{15-25} \cline{27-37}

\multicolumn{37}{c}{} \\[-3ex] \cline{1-1} \cline{3-13} \cline{15-25} \cline{27-37}

9  &  & x & x & x & x & x & x & x & x & x & 1.2 & 1.2   &  & x & x & x & x & x & x & x & x & x & 1.9 & 4.9   &  & x & x & x & x & x & x & x & x & x & 0.3 & 3.6  \\ \cline{1-1} \cline{3-13} \cline{15-25} \cline{27-37}

\multicolumn{13}{c}{} \\

\end{tabular}

\end{center}
\caption{VSSIRIS database: Verification results for an
increasing number of fused comparators. The best combinations are chosen
based on the lowest FRR @ FAR=0.01\% of cross-sensor experiments.
The best result of each column is marked in bold. 
} \label{tab:vssiris-results-fusion}
\end{table}
\normalsize

\setlength{\tabcolsep}{6pt}

\begin{figure}[htb]
\centering
        \includegraphics[width=0.95\textwidth]{FAFR_all_together_vssiris_shortname.png}
\caption{VSSIRIS database: cross-sensor FA/FR curves of the individual systems (left: with raw scores, middle: after z-score normalization, right: after mapping to log-likelihood ratios). Solid curves represent FR curves, while dashed curves represent FA curves. The 'fusion' curves on the center and right plots represent the fusion of SAFE+SIFT+LBP+Densenet201 (see the main text for details). Best seen in colour and zoomed.
}
\label{fig:vssiris-results-scores-FAFR}
\end{figure}

Similarly as Cross-Eyed, cross-sensor performance is also improved
significantly here by fusion. 
The relative EER and FRR improvement of the best fusion case is even bigger,
being 87.5\% and 95.2\%, respectively.
This is high in comparison with the reductions observed with
Cross-Eyed, which were in the order of 30-40\%.
It is also remarkable that similar or even better absolute performance values are obtained with VSSIRIS.
This is despite the worse performance observed in the individual comparators, as discussed in the previous section.
However, it comes at the price of needing more comparators to achieve maximum performance.
Even if the biggest performance improvement also occurs after the fusion of two or three comparators,
the smallest error is obtained with the fusion of four comparators.
In contraposition, Cross-Eyed needed only two or three (see Figure~\ref{fig:crosseyed-results-fusion}).

The fusion methods evaluated also rank in the same order here (see Figure~\ref{fig:vssiris-results-fusion}).
The probabilistic fusion method based on calibration (LLR) outperforms all the others,
followed by SVM linear and polynomial. The simple average rule also matches the
performance of other trained approaches in some points, but it deteriorates quickly
as more comparators are combined. Lastly, the Random Forest approach performs the worst in general.
In addition, the SIFT comparator is also decisive to achieve lower error rates,
as it is always selected in any combination (Table~\ref{tab:vssiris-results-fusion}).
The CNN comparators are also selected first, but to achieve the best performance,
the role of other comparators are decisive with this database.
The best FRR, for example, is given by the combination of SAFE, SIFT, LBP and DenseNet201.
The same can be said with other fusion methods. The best FRR with the average fusion
involves SIFT, NTNU, and DenseNet201, while the best FRR with the linear SVM engages
SAFE, GABOR, SIFT, HOG and DenseNet201.

Figure~\ref{fig:vssiris-results-scores-FAFR} provides the FA/FR curves of the systems with different score normalizations. A selected fusion case is also plotted (SAFE+SIFT+LBP+DenseNet201, best combination of four systems in Table~\ref{tab:vssiris-results-fusion}). The same observations than Section~\ref{sect:crosseyed-fusion} can be made, in the sense that calibration provides alignment of genuine and impostor distribution around zero, and that the arrangement and spread of the distributions to both sides of the horizontal axis are indicative of the relative performance among systems.

\section{Conclusion}
\label{sect:conclusion}

Periocular biometrics has rapidly evolved to competing with face or
iris recognition \cite{[Alonso16],[Nigam15]}. The periocular region has shown
to be as discriminative as the full face, with the advantage that it
is more tolerant to variability in expression, blur, downsampling
\cite{[Miller10]}, or occlusions \cite{[Juefei-Xu14],[Park11]}.
Under difficult conditions, such as people walking by acquisition portals,
\cite{[Woodard10a],[Boddeti11],[Ross12]}, distant acquisition,
\cite{[Tan12],[Tome13_TIFS_SoftBiometrics]},
smartphones, \cite{[Santos14]}, webcams, or digital
cameras, \cite{[Alonso15],[Alonso15a]}, the periocular modality is
also shown to be clearly superior to the iris modality, mostly due
to the small size of the iris or the use of visible illumination.
The COVID-19 pandemic has also imposed the necessity of developing technologies capable of dealing with faces occluded by protective face masks, often
with just the periocular area visible \cite{[btt20covidFPshut],[Ngan20NISTmasksreport],[Klare20rankonemasks]}.

As biometric technologies are extensively deployed, it will be
common to compare data captured with different sensors or from
uncontrolled non-homogeneous environments. Unfortunately, the comparison
of heterogeneous biometric data for recognition purposes is known to
decrease performance significantly \cite{[Jain16]}.
Hence, as new practical applications evolve, new challenges arise,
as well as the need for developing new algorithms to address them.
In this context, we address in this paper the problem of biometric
sensor interoperability, with recognition by periocular images as
test-bed.

Inspired by our submission to the 1$^{st}$ Cross-Spectral Iris/Periocular Competition
(Cross-Eyed) \cite{[sequeira16crosseyed]},
we propose to mitigate such problem via a multialgorithm fusion strategy at the score level
that combines up to nine different periocular comparators.
The aim of this competition was to evaluate periocular
recognition algorithms when images from visible and near-infrared
spectra are compared.
We follow a probabilistic score fusion approach
based on linear logistic regression \cite{brummer06,[brummer07fusion]}.
With this method, scores from multiple comparators are fused together
not only to improve the discriminating ability
but also to produce log-likelihood ratios as output scores.
%
%
This way, output scores are always in a comparable probabilistic domain
since log-likelihood ratios can be interpreted
as a degree of support to the \textit{target} or \textit{non-target} hypotheses.
This allows the use
of Bayes thresholds for optimal decision-making, avoiding the need to
compute comparator-specific thresholds.
This is essential in operational conditions since the threshold is
critical to determine the accuracy of the
authentication process in many applications.
In the experiments of this paper, this method is shown to surpass other
fusion approaches such as
the simple arithmetic average of normalized scores \cite{[Jain05]}
or trained algorithms such as
Support Vector Machines \cite{[Gutschoven00svmfusion]}
or Random Forest \cite{[Ma05rffusion]}.
%
%
This employed fusion approach has been applied previously to cross-sensor comparison
of face or fingerprint modalities \cite{[Alonso10]} as well,
also providing excellent results in other competition benchmarks
involving these modalities \cite{[Poh09]}.
We employ in this paper three different comparators based on the most widely used
features in periocular research \cite{[Park11]},
as well as three in-house comparators that we proposed recently
\cite{[Alonso16a],[Alonso15],[Raja17]},
and three comparators
based on deep Convolutional Neural Networks \cite{[Parkhi15],[He16],[Huang17]}.
The proposed fusion method,
with a subset of the periocular comparators employed here,
was used in our submission to the mentioned Cross-Eyed evaluation,
obtained the first position in the ranking of participants.
This paper is complemented with
cross-sensor periocular experiments using images from the same
spectrum as well.
For this purpose, we use the Visible Spectrum Smartphone Iris database (VSSIRIS)
\cite{[Raja14b]}, which contains images in the visible range
from two different smartphones.

We first analyze the individual comparators employed not only from the
point of view of its cross-sensor performance (Figures~\ref{fig:crosseyed-results1} and \ref{fig:vssiris-results1}), but also taking into account
its template size and computation times (Tables~\ref{tab:vector-size} and \ref{tab:file-size-time}).
We observe that the comparator having the biggest template size
and computation time is usually the most accurate in terms of individual
performance, also contributing decisively to the fusion.
In the experiments reported in this paper, significant improvements in
performance are obtained with the proposed fusion approach,
leading to an
EER of 0.2\% in visible-to-near-infrared comparisons (Figure~\ref{fig:crosseyed-results-fusion}) and
0.3\% in visible-to-visible comparison of smartphone images (Figure~\ref{fig:vssiris-results-fusion}).
The FRR in high-security environments (at FAR=0.01\%) is also very good,
being 0.47\% and 0.3\%, respectively.
%
%

Interestingly, the best performance is not obtained necessarily by the
combination of all available comparators. Instead, the best results
are obtained by fusion of just two to four comparators.
A fundamental problem in classifier combination is to determine which systems to retain in order to attain the best results \cite{[RaudysJain91smallsampleclassifiers]}.
%
%
The systems retained are not necessarily the best individual ones, especially if they are not sufficiently complementary (for example, if they employ similar features) \cite{[Fierrez05d]}.
When the comparators are properly chosen (in our case, found by exhaustive search), the performance increases quickly with the addition of a small number of them. Then, it tends to stabilize until the addition of new ones actually decreases the performance. The need to retain the best features only, and the mentioned performance `peaking' effect, is well documented \cite{[RaudysJain91smallsampleclassifiers]}, and it can be attributed to the correlation between classifiers or to the effect of a limited sample size.
Such phenomenon have been also observed in other related studies in biometrics \cite{[Roli02MCScomparison],[Fierrez05d],[Alonso07],[Garcia-Salicetti06],[Alonso15a]}.
It is also worth noting that the comparators
producing the best fusion performance (Tables~\ref{tab:crosseyed-results-fusion} and \ref{tab:vssiris-results-fusion})
have an individual performance that differs in
one or two orders of magnitude in some cases.
%
%
In the probabilistic approach employed, each comparator is implicitly
weighted by its individual accuracy, so the most reliable
ones will have a dominant role \cite{[Fierrez05b]}.
It is, therefore, a very efficient method to cope with
comparators having heterogeneous performance.
On the contrary, in conventional
score-level fusion approaches (like the average of scores), each comparator
is given the same weight regardless of its accuracy,
a common drawback that makes
the worst comparators to produce misleading results
more frequently \cite{[Jain05]}.
Another relevant observation is that cross-sensor error rates
of the individual comparators
are higher with the database captured in the same spectrum (VSSIRIS)
than the database which contains images in different spectra (Cross-Eyed).
As a result, there is a need to fuse more comparators with VSSIRIS to achieve
maximum performance.
This is an interesting phenomenon since one would expect that
the comparison of images captured with visible cameras would
produce better results
than the comparison of near-infrared and visible images.
Some authors point out that the discrepancy in colours between
sensors in the visible range can be very important,
leading to a significant decrease in performance when
images from these sensors are compared without applying
appropriate device-dependent colour corrections \cite{[Santos14]}.
Since NIR images do not contain colour information, this effect may not appear in NIR-VIS comparisons.

In the present work,
we use the eye corners or the sclera boundary as references
to extract the periocular region of interest (ROI).
%
%
While we have employed ground-truth information,
an operational system would demand to locate these parts,
so inaccuracies in their location would affect subsequent processing steps.
In order to mitigate the effects of incorrect detection on the periocular matching performance of the different comparators and obtain a measure of their capabilities in ideal conditions \cite{[Park11]},
we have not implemented any detector of the necessary references.
Even if errors in the detection will influence
the overall performance of the recognition chain,
feature extraction methods are not necessarily affected in the same way.
This is seen for example in \cite{[Hofbauer16iet]} with the iris modality,
which will serve as inspiration for a similar systematic study with periocular images.
The amount of periocular area around the eye necessary to provide good accuracy is another subject of study, with studies showing differences depending on the spectrum \cite{[Alonso14b]}. In VSSIRIS, the available images (captured with smartphones) contain a bigger periocular portion than images from the Cross-Eyed database (Figure~\ref{fig:db-samples}). However, it is not sufficient to provide better \emph{cross-sensor} accuracy. Therefore, an interesting source of future research work will be to test the resilience against a variable amount of periocular area, including occlusions \cite{[Park11]}.
%

Another observation is that the proposed fusion method needs to be trained separately for each domain (NIR-VIS or VIS-VIS).
This is not exclusive of this method but an issue that is common to score-level fusion methods in general.
Since the scores given by different systems do not necessarily lie in the same range, they are made comparable by mapping them to a common interval using score normalization techniques \cite{[Fierrez18]}.
Even the score distributions of a given algorithm do not necessarily lie in the same range if the operational conditions are different, such as operating in NIR-VIS or VIS-VIS domains. Just changing a sensor by a more recent one from the same manufacturer 
may have the same effect \cite{[Alonso17b_eusipco_busch]},
and the shape of the distributions are not necessarily equal either.
%
%
One obvious effect of the difference between score distributions in different domains is that the accuracy of the comparators is different, not only in absolute numbers but also in the relative differences among them (Table~\ref{tab:crosseyed-results} vs Table~\ref{tab:vssiris-results}). For example, the best comparator in Table~\ref{tab:crosseyed-results} is SIFT, and it is one order of magnitude better than the others. On the other hand, in Table~\ref{tab:vssiris-results}, the EER of SIFT is only a little ahead of Resnet101 or Densenet201, and the FRR is even worse.
Another observable effect of this phenomenon is that
the slope of the DET curves is not the same either 
(Figure~\ref{fig:crosseyed-results1} vs 
Figure~\ref{fig:vssiris-results1}).  
For these reasons, the normalization and the fusion algorithms will usually need different training for each context. The calibration method employed implicitly finds the weight to be given to each system, so if their absolute or relative performance changes, the weights need to change accordingly. The same can be said about the other fusion algorithms evaluated.
The number of systems that are needed to achieve maximum performance will not necessarily be the same either (Figure~\ref{fig:crosseyed-results-fusion} vs Figure~\ref{fig:vssiris-results-fusion}), nor the individual systems involved in the fusion (Table~\ref{tab:crosseyed-results-fusion} vs Table~\ref{tab:vssiris-results-fusion}).
These observations are also backed up by a number of previous studies with different biometrics modalities \cite{[Alonso15a],[Tome10],[Alonso09c],[Fierrez05d]}.
As a future work in this direction, 
we are looking at the robustness of the different comparators to cross-domain training, i.e. training the calibration in one domain and testing in the other. We speculate that some comparators may be more robust than others, so using only those for calibration would allow transferring the training for one domain to the other without needing to re-train in the target domain.
%
%
The use of several databases in one domain is also another way to test the generalization of the suggested approach by cross-database training \cite{[Lopez19iet_face_dbbias]}.
%
%
%
As future work, we are also exploring to exploit deep learning frameworks
to learn the variability between images in different spectra or
captured with different sensors.
One plausible approach is the use of
Generative Adversarial Networks \cite{[Goodfellow14]}
to map images as if they were captured by the same sensor.
This has the advantage that images can be compared
using standard feature extraction methods such as the ones
employed in this paper, which have been shown
to work better if images are captured using the same sensor.
%
%

In the context of smartphone recognition, where high-resolution
images may be available, fusion with the iris modality is another
possibility to increase recognition accuracy \cite{[Alonso15a]}.
However, it demands segmentation,
which might be an issue if the image quality is not sufficiently
high \cite{[Alonso12a]}.
This motivates pursuing the periocular modality, as in
the current study.
We will also validate our methodology using databases not only
limited to two devices or spectra, e.g. \cite{[Santos14],[Vetrekar18]},
and also including
more extreme variations in camera specifications and imaging conditions,
such as low resolution, illumination or pose variability.
For such low-quality imaging conditions, super-resolution techniques
may also be helpful \cite{[Alonso19_iris_SR_IEEEaccess]} and will be investigated as well.

Finally, recent interest in learning biases around face recognition \cite{Drozdowski20TTS_DemographicBiasSurvey,terhorst21_StudyFaceBiasBeyondDemographics} motivates future research to study learning biases in the periocular region and developing new methods to reduce undesired biases \cite{[Morales21SensitiveNets]} in that important facial region.

\section*{Acknowledgment}

Part of this work was done while F. A.-F. was a visiting
researcher at the Norwegian University of Science and Technology in
Gj\o{}vik (Norway), funded by EU COST Action IC1106.
Authors from HH thank the Swedish Research Council (project 2016-03497),
the Swedish Knowledge Foundation (CAISR and SIDUS-AIR Program), and the
Swedish Innovation Agency VINNOVA (project 2018-00472)
for funding his research.
%
%
Authors from UAM are funded by projects: PRIMA (MSCA-ITN-2019-860315), TRESPASS-ETN (MSCA-ITN-2019-860813), and BIBECA (RTI2018-101248-B-I00 MINECO).



\end{document}